\def\year{2022}\relax
\documentclass[letterpaper]{article} 
\usepackage{aaai22}  
\usepackage{times}  
\usepackage{helvet}  
\usepackage{courier}  
\usepackage[hyphens]{url}  
\usepackage{graphicx} 
\usepackage{natbib}  
\usepackage{caption} 
\DeclareCaptionStyle{ruled}{labelfont=normalfont,labelsep=colon,strut=off} 
\usepackage{algorithm}
\usepackage{algorithmic}

\usepackage{newfloat}
\usepackage{listings}
\floatstyle{ruled}
\newfloat{listing}{tb}{lst}{}
\floatname{listing}{Listing}
\usepackage{booktabs}
\usepackage{multirow}

\usepackage{amsmath,amsfonts,bm}

\newcommand{\eqdef}{\overset{def}{=}}

\def\eqref#1{equation~\ref{#1}}

\def\1{\bm{1}}

\DeclareMathAlphabet{\mathsfit}{\encodingdefault}{\sfdefault}{m}{sl}
\SetMathAlphabet{\mathsfit}{bold}{\encodingdefault}{\sfdefault}{bx}{n}

\DeclareMathOperator*{\argmin}{arg\,min}

\title{Improving Evidential Deep Learning via Multi-task Learning}
\author{%
  Dongpin Oh, \textsuperscript{\rm 1} Bonggun Shin \textsuperscript{\rm 2}\\
}
\affiliations{
    \textsuperscript{\rm 1} Deargen Inc., Seoul, South Korea\\
    \textsuperscript{\rm 2} Deargen USA Inc., Atlanta, GA
}

\begin{document}

\frenchspacing
\maketitle
\begin{abstract}
   The Evidential regression network (ENet) estimates a continuous target and its predictive uncertainty without costly Bayesian model averaging. However, it is possible that the target is inaccurately predicted due to the gradient shrinkage problem of the original loss function of the ENet, the negative log marginal likelihood (NLL) loss. In this paper, the objective is to improve the prediction accuracy of the ENet while maintaining its efficient uncertainty estimation by resolving the gradient shrinkage problem. A multi-task learning (MTL) framework, referred to as MT-ENet, is proposed to accomplish this aim. In the MTL, we define the Lipschitz modified mean squared error (MSE) loss function as another loss and add it to the existing NLL loss. The Lipschitz modified MSE loss is designed to mitigate the gradient conflict with the NLL loss by dynamically adjusting its Lipschitz constant. By doing so, the Lipschitz MSE loss does not disturb the uncertainty estimation of the NLL loss. The MT-ENet enhances the predictive accuracy of the ENet without losing uncertainty estimation capability on the synthetic dataset and real-world benchmarks, including drug-target affinity (DTA) regression. Furthermore, the MT-ENet shows remarkable calibration and out-of-distribution detection capability on the DTA benchmarks.
\end{abstract}

\section{Introduction}

The essential task in the safety-critical deep learning systems is to quantify uncertainty \cite{gal2016uncertainty, kendalluncertainties}. The factors contributing to uncertainty can be classified into two types: irreducible observation noise (\textit{aleatoric uncertainty}) and the uncertainty of model parameters (\textit{epistemic uncertainty}) \cite{gal2016uncertainty, guo2017calibration}. In particular, the difficulty and expense involved in representing the uncertainty of the model parameters make it challenging to quantify epistemic uncertainty. 

Common approaches used to estimate epistemic uncertainty are Ensemble-based methods and Bayesian neural networks (BNNs) \cite{wilson2020bayesian}. Ensemble-based methods and BNNs have been shown to produce impressive results in terms of both accuracy and the robustness of uncertainty estimation \cite{welling2011bayesian, neal2012bayesian, blundell2015weight, lakshminarayanan2017simple, gal2016dropout, maddox2019simple}. However, since multiple numbers of models are required in ensemble models, and expansive approximations are necessary in BNNs for the intractable posterior, neither of these methodologies can be considered cost-effective in real-world applications \citep{amini2020deep}.

In contrast, the evidential neural network (ENet) is a cost-effective \textit{deterministic} neural network, designed to accomplish uncertainty estimation by generating conjugate prior parameters as its outputs \citep{malinin2018predictive, sensoy2018evidential,gurevich2020gradient, malinin2020regression}. The ENet can model both epistemic and aleatoric uncertainty based on the uncertainty of the generated conjugate prior. The ability of the ENet to estimate uncertainty without an ensemble or an expensive posterior approximation is remarkable.

On the other hand, the fundamental goal of deep learning is to accomplish state-of-the-art predictive accuracy, not only to estimate uncertainty. Although the ENet architecture can achieve outstanding and practical uncertainty estimations, NLL loss (negative log marginal likelihood)---the original loss function of the ENet---may result in the high mean squared error (MSE) of the ENet's prediction in a certain condition. Intuitively speaking, this problem is caused by the fact that the NLL loss could finish training by determining that the target values are unknown instead of correcting the prediction of the ENet. We show that this phenomenon occurs when the estimated epistemic uncertainty is high. This high epistemic uncertainty implies that the given training samples are sparse or biased \citep{gal2016uncertainty}, which is a common situation in real-world applications, such as drug-target affinity prediction tasks \citep{ezzat2016drug, yang2021delving}. 

One possible solution to resolve this issue is reformulating the training objective of the ENet as the Multi-task learning (MTL) loss with an additional loss function which only optimizes the predictive accuracy, such as the MSE loss. However, in the MTL, the conflicting gradients problem could occur, which negatively impact performance \citep{sener2018multi, lin2019pareto, yu2020gradient}. In particular, for the safety-critical systems, it can be harmful if the uncertainty estimation capability is degraded despite improved accuracy due to the gradient conflict between the losses. Therefore, to identify and determine the reason for the gradient conflict in our MTL optimization, the gradient between the loss functions was analyzed. Our analysis shows the certain condition that this gradient conflict could occur.

Based on this analysis, we define the Lipschitz modified MSE loss, which is designed to mitigate the gradient conflict with the original NLL loss function of the ENet. We also propose an MTL framework using the Lipschitz MSE, named Multi-task-based evidential network (MT-ENet). In particular, our contributions are as follows:

\begin{itemize}
    \item We thoroughly show that (1) the NLL loss alone is not sufficient to optimize the prediction accuracy of the ENet, (2) and adding the MSE-based loss into the training objective can resolve this issue.
    \item We establish the condition that the MSE-based loss function could conflict with the NLL loss. To avoid this gradient condition, the Lipschitz modified MSE loss function is designed. Based on our novel Lipschitz loss function, we propose an MTL framework, MT-ENet, that improves the prediction accuracy of the ENet while maintaining its uncertainty estimation capability.
    \item Our experiments show that MT-ENet outperforms other strong baselines on the real-word regression benchmark datasets. And the MT-ENet shows remarkable performance in uncertainty estimation and out-of-distribution detection on the drug-target affinity datasets.
\end{itemize}

\section{Background}
\label{sec:2}
\subsection{Problem setup} 
Assume we have a dataset of the regression task,  $\mathcal{D}=\{(X_i,y_i)\}^N_{i=1}$, where $X_i \in \mathbb{R}^d$ are the i.i.d. input data points, $d$ is the dimension of an input vector, $y_i \in \mathbb{R}$ are the real-valued targets, and $N$ is the number of data samples. We consider resolving a regression task by modeling a predictive distribution, $p(y|f_\theta(X))$, where $f$ is a neural network and $\theta$ is its parameters.

\begin{figure}[t]
\includegraphics[width=\columnwidth]{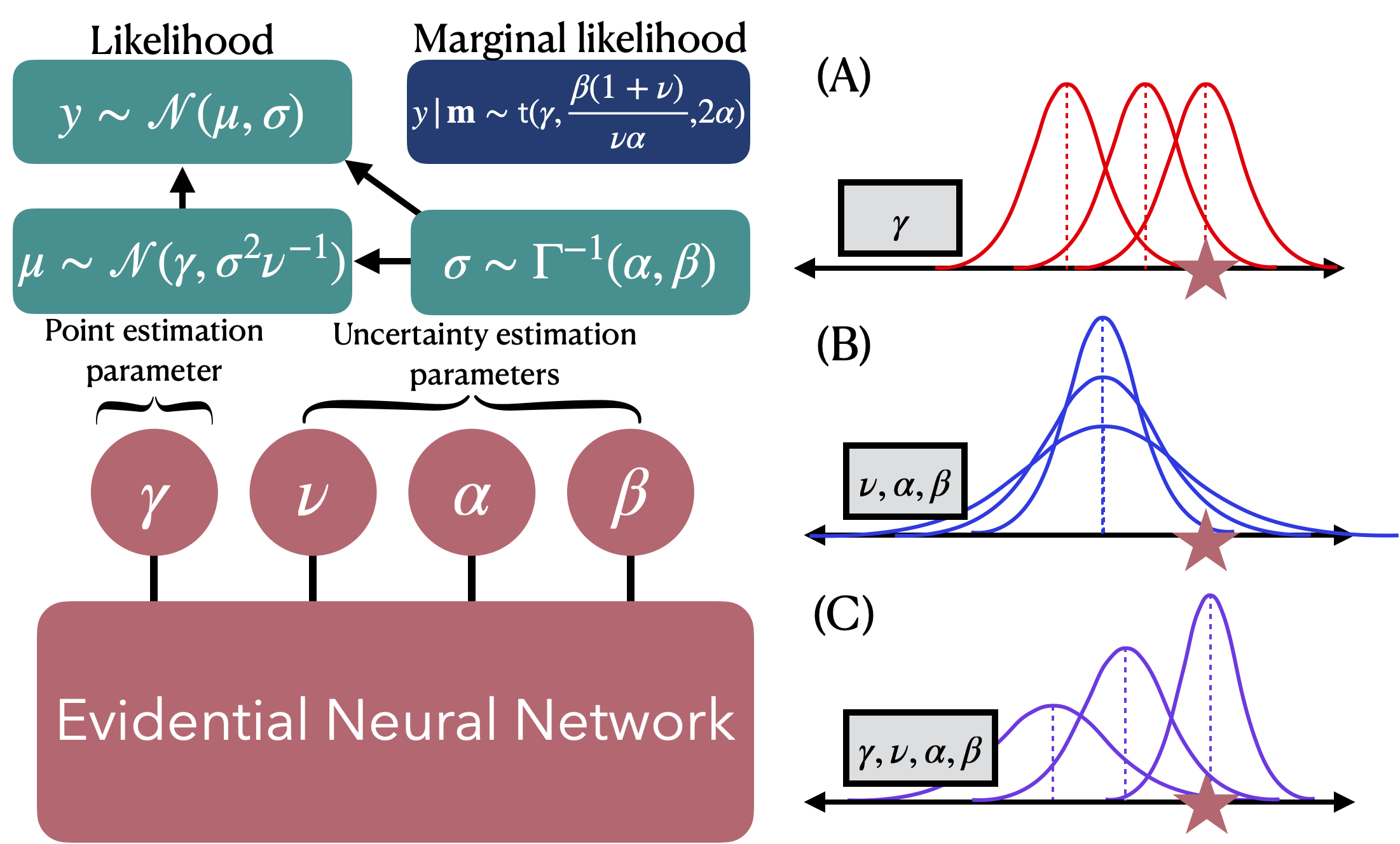}
\caption{A scheme of the architecture of the MT-ENet and ENet and simplified examples of roles of the outputs ($\mathbf{m}=\{\gamma,\nu,\alpha,\beta\}$). (Left) An overview of the ENet architecture. The likelihood and marginal likelihood distribution can be calculated via the outputs of the ENet. (A) $\gamma$ determines the model prediction (point estimation); (B)  $\nu, \alpha,\beta$ determine the model uncertainty (uncertainty estimation); (C) The predictive distribution (Marginal likelihood) of the ENet can be derived by $\gamma,\nu,\alpha,\beta$.}
\label{fig:fig1}
\end{figure}

\subsection{Evidential regression network} 
An evidential regression network (ENet) \citep{amini2020deep} considers a target value, $y$, as a sample drawn from a normal distribution with unknown parameters, {$\mu, \sigma$}. These parameters, $\mu$ and $\sigma$, are drawn from a Normal-Inverse-Gamma (NIG) distribution, which is the conjugate prior to the normal distribution:
\begin{equation}
\begin{gathered}
\label{eq1}
y \sim \mathcal{N}(\mu,\sigma^2),\quad \mu \sim \mathcal{N}(\gamma, \tfrac{\sigma^2}{\nu}),\quad  \sigma^2\sim \text{Gamma}^{-1}(\alpha,\beta) 
\end{gathered}
\end{equation}
where $\gamma \in \mathbb{R}, \nu > 0, \alpha > 1, \beta > 0$, and $\text{Gamma}^{-1}$ is the inverse-gamma distribution. The NIG distribution in Eq~\ref{eq1}. is parameterized by $\mathbf m = \{\gamma, \nu, \alpha, \beta\}$, which is \textit{the output of the ENet}, $\mathbf{m}=f_\theta(X)$, where $\theta$ is a trainable parameter of the ENet (Fig~\ref{fig:fig1}).

With the NIG distribution in Eq~\ref{eq1}, the model prediction ($\mathbb{E}[\mu]$), aleatoric ($\mathbb{E}[\sigma^2]$), and epistemic ($Var[\mu]$) uncertainty of the ENet can be calculated by the following:
\begin{equation}
\label{eq2}
\mathbb{E}[\mu] = \gamma, \quad \mathbb{E}[\sigma^2] = \frac{\beta}{\alpha - 1}, \quad  Var[\mu] = \frac{\beta}{\nu(\alpha - 1)}.
\end{equation}
With these equations, we define \textbf{point estimation} as estimating the model prediction of the ENet, $\mathbb{E}[\mu] = \gamma$ and  \textbf{uncertainty estimation} as estimating the uncertainties, $\mathbb{E}[\sigma^2]$ and $Var[\mu]$. 

In addition to the uncertainty estimates in Eq~\ref{eq2}, \textbf{pseudo observation} can be used as an alternative interpretation of the predictive uncertainty, $\mathbb{E}[\sigma^2]$ and $Var[\mu]$ \citep{murphy2007conjugate,murphy2012machine}. This is widely used in Bayesian statistics literatures \citep{lee1989bayesian}.

\paragraph{Definition 1.} \label{def2}
\textit{We define $\alpha$ and $\nu$, the outputs of the ENet, as the \textbf{pseudo observations}.}\\
Note that the aleatoric and the epistemic uncertainty increase as the pseudo observations decrease because they are inversely proportional ($\mathbb{E}[\sigma^2] \propto \tfrac{1}{\alpha}$, $Var[\mu] \propto \tfrac{1}{\alpha \nu}$).

\subsection{Training the ENet with the marginal likelihood}
The ENet learns its trainable parameters $\theta$ by maximizing a \textit{marginal likelihood} with respect to unknown Gaussian parameters, $\mu$ and $\sigma$. By analytically marginalizing the NIG distribution (Eq~\ref{eq1}) over these parameters $\mu$ and $\sigma$, we can derive the following marginal likelihood: 
\begin{equation}
\begin{aligned}
\label{eq3}
p(y|\mathbf m) &= \int_{\sigma^2=0}^{\sigma^2=\infty}\int_{\mu=-\infty}^{\mu=\infty} p(y|\mu,\sigma^2)p(\mu,\sigma^2| \mathbf m ) d\mu d\sigma^2\\
&=\text{t} (y; \gamma, \dfrac{\beta(1+\nu)}{\nu\alpha}; 2\alpha)
\end{aligned}
\end{equation}
where t$(x; l, s, n)$ is the student-t distribution with the location parameter ($l$), the scale parameter ($s$), and the degrees of freedom ($n$).

The training procedure for the ENet is to minimize the negative log marginal likelihood (NLL) loss function, $L_{NLL}(y,\mathbf{m})=- \log(p(y|\mathbf m))$, as summarized in the following equation:
\begin{equation}
\begin{split}
\label{eq4}
\argmin_\theta &(L_{NLL}(y,\mathbf{m})) = \argmin_\theta \tfrac{1}{2}\log (\tfrac{\pi}{\nu}) - \alpha \log \Lambda\\
&+ (\alpha + \tfrac{1}{2})\log ((y - \gamma)^2\nu + \Lambda) + \log(\tfrac{\Gamma(\alpha)}{\Gamma(\alpha + \tfrac{1}{2})})
\end{split}
\end{equation}
where $\Gamma(\cdot)$ is the gamma function and $\Lambda = 2\beta(1+\nu)$. Besides allowing the ENet to predict a proper point estimate $\gamma$, the NLL loss function allows the ENet to quantify the aleatoric and epistemic uncertainty via Eq 4.

\section{Gradient shrinking of the NLL loss}
\label{sec:3}
In this section we show that the NLL loss alone is not sufficient to optimize the prediction accuracy of the ENet. Even though the model trained on NLL loss can predict the proper point estimate, $\gamma$, the model can circumvent to achieve lower NLL loss by increasing the predictive uncertainty \textbf{instead of achieving a higher predictive accuracy}. This limitation can be resolved by using an additional loss function. To formally state this, we first define the gradient vector of the loss function:
\paragraph{Definition 2.} \label{def4} \textit{Consider a loss function $L:\mathbb{R}^n \rightarrow \mathbb{R}$ and a set of outputs of a model, $f_\theta(X) \in\mathbb{R}^n$, with parameters $\theta$. Let $\mathbf{\Omega}_\theta$ be the subset of the model outputs, $\mathbf{\Omega}_\theta \subseteq f_\theta(X)$, then $\mathbf{g}_{L,\mathbf{\Omega}} = \sum_{\omega_\theta \in \Omega_\theta} \tfrac{\partial L}{\partial\mathbf{\omega_\theta}} \nabla_\theta \mathbf{\omega_\theta}$ denotes the \textbf{gradient vector} with respect to the subset of the model outputs, $\mathbf{\Omega}_\theta$. We exclude the reliance of $\theta$ to simplify notations.}\\

In order to increase the accuracy of the ENet, the model prediction ($\gamma$) should be trained to get closer to the true value via the corresponding gradient $\mathbf{g}_{L_{NLL},\gamma}$. For the ideal loss function, its gradient magnitude becomes zero only when the model prediction and the true value are identical. However, \textbf{despite an inaccurate prediction} if the prediction uncertainty is severely increased by the NLL loss, the gradient for the model prediction ($\mathbf{g}_{L_{NLL},\gamma}$) could be significantly small. Thus, we aim to verify that the gradient magnitude for the model prediction, $\|\mathbf{g}_{L_{NLL},\gamma}\|$, is insignificant despite of the incorrect model prediction $\gamma$. Specifically, we prove that the $\|\mathbf{g}_{L_{NLL},\gamma}\|$ is converged to zero under the condition that the pseudo observation ($\nu$) becomes zero as shown in \textit{Theorem 1}:

\paragraph{Theorem 1.} \label{them1} (Shrinking NLL gradient)\footnote{Proofs and details of all the theorems and propositions of this paper are given in Appendix A.}. \textit{Let $\nu>0, \gamma\in\mathbb{R}$ be the output of the ENet, and assume that $(y-\gamma)^2>0$, then for every real $\epsilon > 0$, there exists a real $\delta > 0 $ such that for every $\nu$, $0 < |\nu| < \delta$ implies $|\tfrac{\partial L_{NLL}}{\partial\gamma}| < \epsilon$. Therefore, $\lim_{\nu \rightarrow 0+}\tfrac{\partial}{\partial\gamma}L_{NLL}(y,\mathbf{m}) = 0$ $\Rightarrow$ $\lim_{\nu \rightarrow 0+}\|\mathbf{g}_{L_{NLL},\gamma}\| = 0$.}\\

\textit{Theorem 1} states that the NLL loss itself is insufficient to correct the point estimate ($\gamma$) because it cannot fully utilize the error value, $(y-\gamma)^2$, when pseudo observation ($v$) is very low. This infinitely low pseudo observation signifies that infinitely high epistemic uncertainty of the ENet ($Var[\mu]$) since $Var[\mu] \propto \tfrac{1}{\nu}$. This statement is empirically justified in the experiment on the synthetic dataset as shown in Fig~\ref{fig:fig3}.


On the other hand, the simple MSE loss, $L_{MSE}(y,\mathbf{m}) = (y-\gamma)^2$, consistently trains the ENet to correctly estimate $\gamma$, since $\tfrac{\partial}{\partial\gamma}L_{MSE}$ is independent of $\nu$. Thus, the MSE-based loss functions---which only updates the target value ($\gamma$)---could improve the point estimation capability of the ENet.

\setlength\intextsep{0pt}
\begin{figure}[t]
  \begin{center}
    \includegraphics[width=\columnwidth]{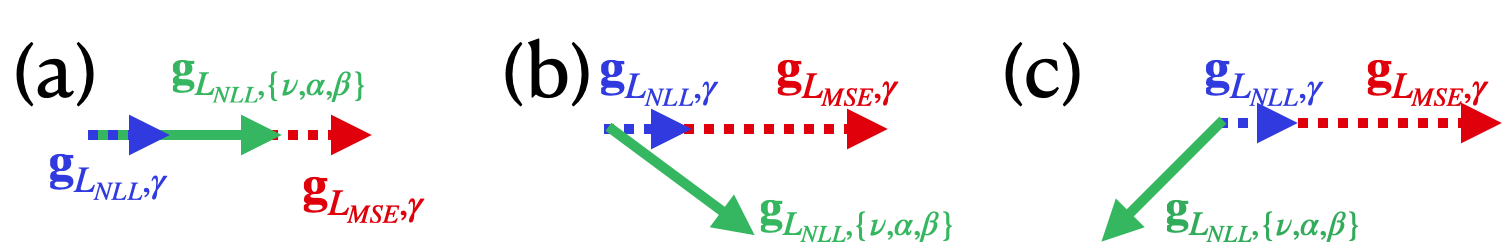}
  \end{center}
  \caption{Simplified examples of the gradient vectors. The red arrows represent {\textbf{the gradient of MSE}}; The blue arrows characterize the {\textbf{point estimation gradient}} of NLL; The green arrows signify the {\textbf{uncertainty estimation gradient}} of NLL. (a) Non-conflicting gradients. (b) Slightly conflicting gradients. (c) Conflicting gradient by $\mathbf{g}_{L_{NLL},\{\nu, \alpha, \beta\}}$. As \textit{Proposition 1} states, the direction of $\mathbf{g}_{L_{MSE},\gamma}$ and $\mathbf{g}_{L_{NLL},\gamma}$ is always the same.}
  \label{fig:fig2}
\end{figure}

\section{Gradient conflict between the NLL and MSE}
\label{sec:4}
The previous section established that the simple MSE loss could resolve the gradient shrinkage problem. This finding motivates us to integrate the MSE loss into the NLL loss to form a MTL framework in order to improve the predictive accuracy. When forming this MTL framework, it is important not to have \textbf{gradient conflict} between tasks because they could lead to performance degradation. In the MTL literature, this undesirable case can be defined as when gradient vectors are in different directions \cite{sener2018multi, yu2020gradient}.

To avoid this degradation, it is neccessary to identify the cause of the gradient conflict. In this section, the gradient conflict between the simple MSE and NLL loss function is considered. To do this, the gradient of NLL is considered by dividing it into two separate gradients: uncertainty estimation gradient and point estimation gradient. Here, the \textit{uncertainty} and \textit{point estimation gradient} of the NLL are clarified with the following definition:

\paragraph{Definition 3.} \textit{We define the \textbf{point estimation gradient} of the NLL as $\mathbf{g}_{L_{NLL},\gamma}$, since $\gamma$ performs the point estimation. The  \textbf{uncertainty estimation gradient} of the NLL is defined as $\mathbf{g}_{L_{NLL},\{\nu,\alpha,\beta\}}$, since $\{\nu,\alpha,\beta\}$ perform the uncertainty estimation. Finally, we define the \textbf{total gradient} of the NLL as $\mathbf{g}_{L_{NLL},\mathbf{m}}$.}\\

Then, we show that the \textit{point estimation gradient of the NLL} from the Definition 3 never conflicts with the gradient of the MSE as shown in the following proposition:
\paragraph{Proposition 1.} \label{them2} (Non-conflicting point estimation)\footnotemark[1]. \textit{If gradient magnitudes of $\mathbf{g}_{L_{MSE},\gamma},\mathbf{g}_{L_{NLL},\gamma}$ are not zero, then the cosine similarity $s(\cdot)$ between $\mathbf{g}_{L_{MSE},\gamma}$, and $\mathbf{g}_{L_{NLL},\gamma}$ is always one:   $s(\mathbf{g}_{L_{MSE},\gamma},\mathbf{g}_{L_{NLL},\gamma}) = 1$}.\\
\\\textit{Proposition 1} states that providing the model updates its parameters only through $\mathbf{g}_{L_{MSE},\gamma}$ and $\mathbf{g}_{L_{NLL},\gamma}$, the two gradients will never conflict. This is indicated by the red arrows and blue arrows in Fig~\ref{fig:fig2}. Then, a trivial consequence of the \textit{Proposition 1} is the following corollary:

\paragraph{Corollary 1.} \textit{If the \textbf{total gradients} of the NLL and the gradient of MSE are not in the same direction, then the \textbf{uncertainty estimation gradients} of the NLL and the gradients of the MSE are also not in the same direction: $s(\mathbf{g}_{L_{MSE},\mathbf{m}}, \mathbf{g}_{L_{NLL},\mathbf{\mathbf{m}}})<1 \Rightarrow s(\mathbf{g}_{L_{MSE},\mathbf{m}}, \mathbf{g}_{L_{NLL},\mathbf{\{\nu,\alpha,\beta\}}}) < 1$.}\\
\\Since \textit{Proposition 1} states that there is no gradient conflict between $\mathbf{g}_{L_{MSE},\gamma}$ and $\mathbf{g}_{L_{MSE},\gamma}$, \textit{Corollary 1} implies that the \textbf{only cause of gradient conflict} between the MSE and NLL loss is the \textbf{uncertainty estimation gradient} of the NLL. Therefore, if the gradient conflict between the uncertainty estimation gradient and the gradient of the MSE loss can be avoided, the MTL can be performed more effectively.

\subsection{Conjecture of the gradient conflict}
\label{sec:4.1}
\setlength\intextsep{0pt}
\begin{figure}[t]
  \begin{center}
    \includegraphics[width=0.47\textwidth]{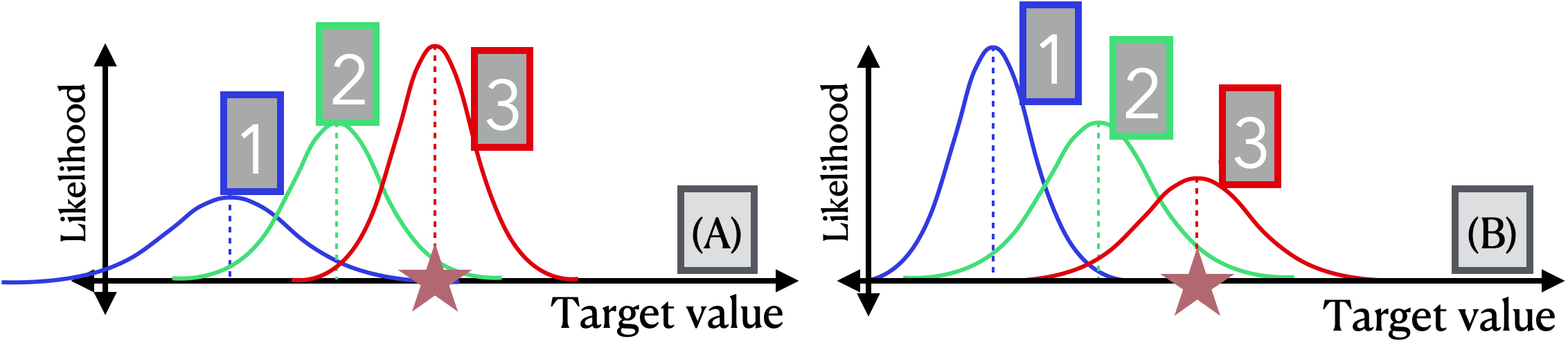}
  \end{center}
  \caption{Illustration of the trained predictive distribution via the MSE and NLL losses. The dashed lines represent the model predictions  ($\mathbb{E}[\mu]=\gamma$); The stars are the ground truth. (A) The NLL decreases the uncertainty; (B) The NLL increases the uncertainty. The optimization procedure is (1)$\rightarrow$(2)$\rightarrow$(3). The NLL loss of (B) is higher than A due to the gradient conflict between the MSE and NLL.}
  \label{fig:illust_nll}
\end{figure}

In the previous section, it was shown that the \textit{uncertainty estimation gradient} of the NLL is the only cause of the gradient conflict. 
In this section, we illustrate and focus on how the increased predictive uncertainty due to uncertainty estimation gradient can result in large gradient conflict. Fig 3 shows the examples of the predictive distributions of the ENet trained on the MSE and NLL loss during its training processes. According to the multi-task learning literature, it is a common premise that the more severe the gradient conflict, the greater the loss \citep{chen2018gradnorm,lin2019pareto,yu2020gradient,chen2020just,javaloy2021rotograd}. In this sense,  since (\textit{B}) has a higher loss than (\textit{A}) as long as the model prediction is approaching its true value, we can consider that the gradient conflict of (\textit{B}) is more significant.

The fundamental difference between (\textit{B}) and (\textit{A}) is that the NLL increases uncertainty in (\textit{B}), and the NLL decreases uncertainty in \textit{(A)}. Therefore, \textbf{we assume that gradient conflicts between the MSE and NLL are more significant when the NLL increases the predictive  uncertainty of the ENet} like the case of (\textit{B}). In practice, to alleviate the gradient conflict of \textit{(B)}, Lipschitz modified MSE loss function is designed. And we empirically demonstrate that our \textit{Lipschitz modified MSE loss function} successfully reduces the gradient conflict as shown in Fig~\ref{fig:conflict}.

\section{Multi-task based evidential neural network}
\label{sec:5}
A new multi-task loss function of the ENet is introduced in this section. The ultimate goal of this multi-task-based evidential neural network (MT-ENet) is to \textit{improve the predictive accuracy} of the ENet \textit{while does not interrupt the uncertainty estimation training on the NLL loss}. To accomplish this, the MT-ENet utilizes a novel Lipschitz loss function as an additional loss function which can mitigate gradient conflicts with the NLL loss.

\subsection{Prevent the conflict via the Lipschitz MSE loss}
\label{sec:5.1}
In the previous section, we established that two gradients from the MSE and NLL could conflict when the NLL increases predictive uncertainty. In this section, this gradient conflict condition is specified to design the loss function that can resolve this conflict.

Specifically, we reinterpret this conflict situation as the change of \textbf{pseudo observations} of the ENet, since an increase in predictive uncertainty is associated with a decrease in the pseudo observation ($\mathbb{E}[\sigma^2] \propto \tfrac{1}{\alpha}$, $Var[\mu] \propto \tfrac{1}{\alpha \nu}$). This decrease in the pseudo observations by the NLL loss occurs when the difference between the model prediction and the true value exceeds specific thresholds, as shown in \textit{Proposition 2}:
\paragraph{Proposition 2.} \label{prop2} \textit{Let $\lambda^2\eqdef(y-\gamma)^2$, which is the squared error value of the ENet. then if $\lambda^2$ is larger than certain thresholds, $U_\nu$ and $U_\alpha$, then the derivative signs of the $L_{NLL}$ w.r.t $\nu$ and $\alpha$ are positive.}
\begin{equation}
\label{eq9}
\begin{gathered}
\lambda^2 >  U_{\alpha} \Rightarrow \frac{\partial}{\partial\alpha} L_{NLL} > 0, \quad
\lambda^2 > U_{\nu} \Rightarrow  \frac{\partial}{\partial\nu} L_{NLL} > 0\\
U_{\nu} = \tfrac{\beta(\nu+1)}{\alpha\nu},\ U_{\alpha} = \tfrac{2\beta(1+\nu)}{\nu}[\exp(\Psi(\alpha + \tfrac{1}{2}) - \Psi(\alpha)) - 1]
\\
\end{gathered}
\end{equation}
\textit{where $\Psi(\cdot)$ is the digamma function.\\}
Since both $\tfrac{\partial L_{NLL}}{\partial\alpha}$ and $\tfrac{\partial L_{NLL}}{\partial\nu}$ contain $\lambda^2$, the thresholds ($U_\nu$ and $U_\alpha$) can be obtained by rearranging $\tfrac{\partial L_{NLL}}{\partial\alpha}$, $\tfrac{\partial L_{NLL}}{\partial\nu}$ to solve for $\lambda^2$ for each case (See Appendix A). 

\textit{Proposition 2} states that when the difference between the predicted and true values is significant, the NLL loss trains the ENet to increase the predictive uncertainty. This increase in predictive uncertainty results in the gradient conflict, which can lead to performance degradation, as explained in the previous section. Thus, our strategy is designing the additional loss function, which regulates its gradient if there is an increase in uncertainty.

\paragraph{Lipschitz MSE loss} We propose a modified Lipschitz MSE loss based on \textit{Proposition 2} to mitigate the gradient conflict that occurs when the MSE is excessively large. Unlike a commonly used MSE loss---which does not have Lipschitzness \cite{qi2020mean}---we define the $L$-Lipschitz continuous loss function where the Lipschitz constant $L$ is dynamically determined by $U_\alpha$ and $U_\nu$ from \textit{Proposition 2}.

Consider a minibatch, $\textbf{X}=\{(X_1,y_1), ..., (X_N,y_N)\}$, extracted from our dataset $\mathcal{D}$ and the corresponding output of the MT-ENet, $f_\theta(X_i)=\mathbf{m}_i=\{\gamma_i, \nu_i, \alpha_i, \beta_i\}$. For each $(X_i,y_i)$ and $\mathbf{m}_i$, the Lipschitz modified MSE $L^*_{mse}$ is defined as follows:
\begin{equation}
\label{eq6}
\begin{gathered}
L^*_{mse}(y_i, \gamma_i) = \begin{cases}
    
    (y_i-\gamma_i)^2, & \text{If }\lambda_i^2 < U_{\nu, \alpha}\\
    
    2\sqrt{U_{\nu, \alpha}}|y_i-\gamma_i| - U_{\nu,\alpha},&\text{If }\lambda_i^2 \geq U_{\nu, \alpha}
    
    \end{cases}
\end{gathered}
\end{equation}
where $U_{\nu,\alpha} \eqdef \min_{i\in [1:N]}(\min(U_{\nu_i},U_{\alpha_i}))$. The $L^*_{mse}$ restricts its gradient magnitude through adjusting its Lipschitz constant, $U_{\alpha,\nu}$, when the pseudo observation tend to be decreased (thus, uncertainty is increased) via the NLL loss. Note that the $L^*_{mse}$ does \textit{not propagate gradients} for the ($\nu, \alpha, \beta$). We empirically demonstrate that the designed Lipschitz modified loss mitigates the gradient conflict with the NLL by illustrating the high cosine similarity between our Lipschitz loss and NLL loss, as shown in Fig 5.

\paragraph{The final objective of the MT-ENet}
The final MTL objective of the MT-ENet, $\mathcal{L}_{total}$, is defined as simple combination of the two losses and the regularization:
\begin{equation}
\label{eq7}
\begin{split}
\mathcal{L}_{total} =& \frac{1}{N}\sum^N_{i=1} L_{NLL}(y_i,\mathbf{m}_i) + L^*_{mse}(y_i,\gamma_i) \\&+ c L_R(y_i,\mathbf{m}_i)
\end{split}
\end{equation}
where $L_R(y, \mathbf{m})=|y-\gamma|(2\nu + \alpha)$ is the evidence regularizer \cite{amini2020deep} and $c > 0$ is its coefficient.

\paragraph{Remark 1.} \textit{The evidential regularization, $L_R$, could also train the model prediction, $\gamma$, like the NLL and MSE. However, the main goal of $L_R$ is not training $\gamma$ but regulating the NLL loss by increasing the predictive uncertainty for incorrect predictions of the ENet \cite{sensoy2018evidential, amini2020deep}. Therefore, $L_R$ cannot play the role of the MSE based loss function, which improves predictive accuracy. Further discussion is available in Appendix B.}

\section{Related work}
\paragraph{Related works on multi-task learning}  Multi-task learning (MTL) is a paradigm for training several tasks with a single model \cite{ruder2017overview}. A naïve strategy of MTL is training a model with a linear combination of given loss functions: $\sum_{i=0}^Tc_iL_i$. However, this linearly combined losses can be ineffective if the loss functions have trade-off relationships between other loss functions \cite{sener2018multi, lin2019pareto}. Several studies have resolved this issue---in the architecture agnostic manner---via dynamically adjusting weights of losses \cite{ sener2018multi, guo2018dynamic, kendall2018multi, liu2019loss, lin2019pareto} or modifying gradients \cite{chen2018gradnorm,chen2020just,yu2020gradient}. Similar to the previous works, we targeted resolving the gradient conflict issue. In particular, we mainly focused on applying the MTL to the ENet while keeping the NLL loss function to maintain the uncertainty estimation capability of the NLL loss function. To accomplish this, we defined the Lipschitz modified MSE which reconciles with the uncertainty estimation by implicitly alleviating the gradient conflict.

\section{Experiments}

\begin{figure}[t]
    \centering
    \includegraphics[width=\columnwidth]{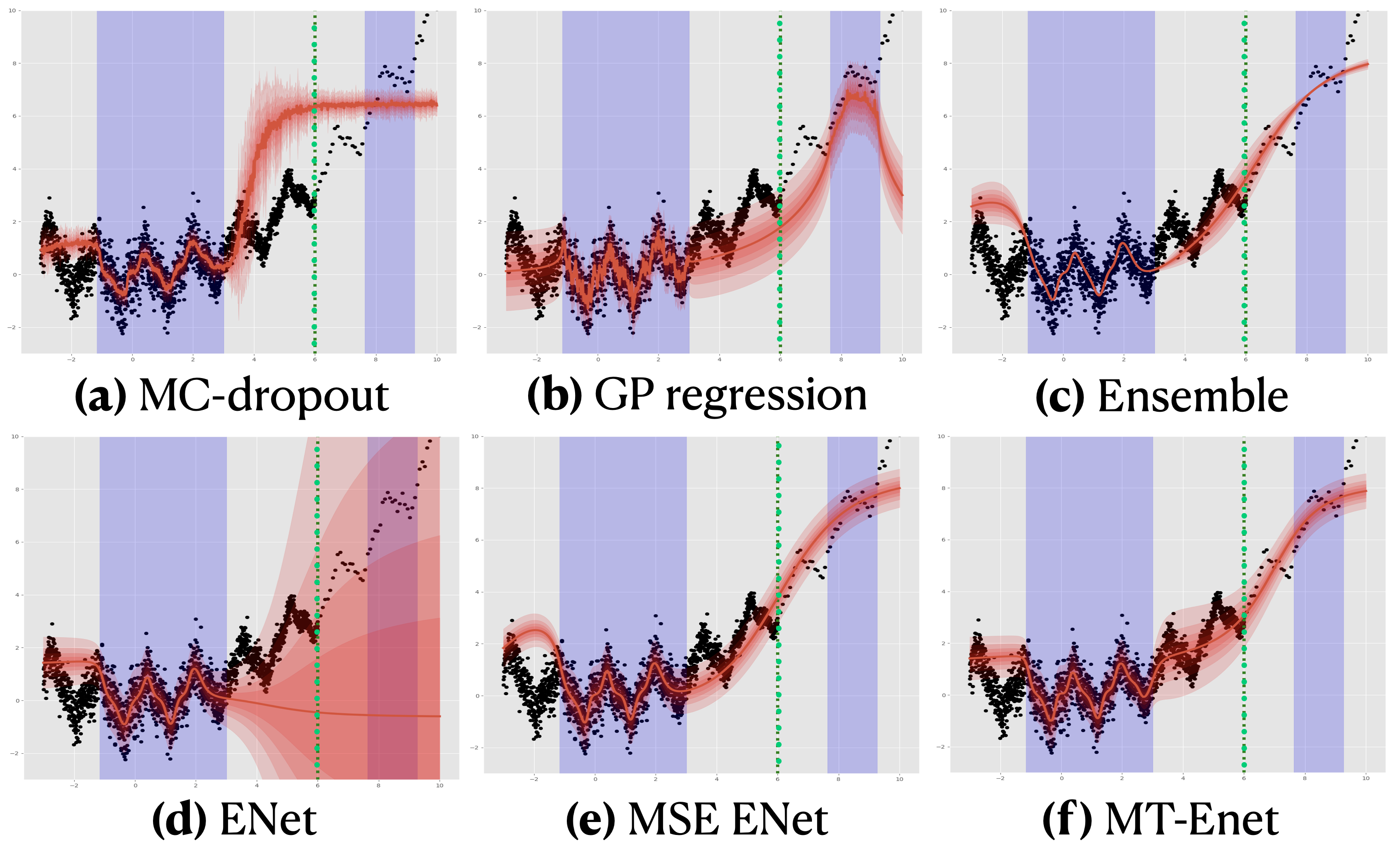}
    \caption{Synthetic dataset results of various models. Red lines represent the predictive mean; Red shades represent the predictive uncertainty; Blue highlighted regions represent our observations. The green lines separate the data-rich and data-sparse regions.}
    \label{fig:fig3}
\end{figure}
\subsection{Synthetic dataset evaluation}
\label{sec:3.1}

\begin{table}[]
\centering
\resizebox{\columnwidth}{!}{\begin{tabular}{@{}lllll@{}}
\toprule
\multicolumn{1}{c}{\multirow{2}{*}{Datasets}} & \multicolumn{4}{c}{\textbf{RMSE}}                                                                 \\ \cmidrule(l){2-5} 
\multicolumn{1}{c}{}                          & MC-Dropout          & Ensemble             & ENet               & MT-ENet               \\ \cmidrule(r){1-5}
\textbf{Boston}                               & \textbf{2.97(0.19)} & 3.28(1.00)           & 3.06(0.16)           & \textbf{3.04(0.21)}  \\
\textbf{Concrete}                             & \textbf{5.23(0.12)} & 6.03(0.58)           & 5.85(0.15)           & \textbf{5.60(0.17)}  \\
\textbf{Energy}                               & \textbf{1.66(0.04)} & 2.09(0.29)           & 2.06(0.10)           & \textbf{2.04(0.07)}  \\
\textbf{Kin8nm}                               & 0.10(0.00)          & \textbf{0.09(0.00)}  & \textbf{0.09(0.00)}  & \textbf{0.08(0.00)}  \\
\textbf{Navel}                                & 0.01(0.00)          & \textbf{0.00(0.00)}  & \textbf{0.00(0.00)}  & \textbf{0.00(0.00)}  \\
\textbf{Power}                                & \textbf{4.02(0.04)} & 4.11(0.17)           & 4.23(0.09)           & \textbf{4.03(0.04)}  \\
\textbf{Protein}                              & \textbf{4.36(0.01)} & 4.71(0.06)           & \textbf{4.64(0.03)}  & 4.73(0.07)           \\
\textbf{Wine}                                 & \textbf{0.62(0.01)} & 0.64(0.04)           & \textbf{0.61(0.02)}  & 0.63(0.01)           \\
\textbf{Yacht}                                & \textbf{1.11(0.09)} & 1.58(0.48)           & 1.57(0.56)           & \textbf{1.03(0.08)}  \\ \toprule
\multicolumn{1}{c}{\multirow{2}{*}{Datasets}} & \multicolumn{4}{c}{\textbf{NLL}}                                                                  \\ \cmidrule(l){2-5} 
                                              & MC-Dropout          & Ensemble             & ENet           & MT-ENet               \\ \cmidrule(r){1-5}
\textbf{Boston}                               & 2.46(0.06)          & 2.41(0.25)           & \textbf{2.35(0.06)}  & \textbf{2.31(0.04)}  \\
\textbf{Concrete}                             & 3.04(0.02)          & 3.06(0.18)           & \textbf{3.01(0.02)}  & \textbf{2.97(0.02)}  \\
\textbf{Energy}                               & 1.99(0.02)          & \textbf{1.38(0.22)}  & 1.39(0.06)           & \textbf{1.17(0.05)}  \\
\textbf{Kin8nm}                               & -0.95(0.01)         & \textbf{-1.20(0.02)} & \textbf{-1.24(0.01)} & -1.19(0.01)          \\
\textbf{Navel}                                & -3.80(0.01)         & -5.63(0.05)          & \textbf{-5.73(0.07)} & \textbf{-5.96(0.03)} \\
\textbf{Power}                                & 2.80(0.01)          & \textbf{2.79(0.04)}  & 2.81(0.07)           & \textbf{2.75(0.01)}  \\
\textbf{Protein}                              & 2.89(0.00)          & 2.83(0.02)           & \textbf{2.63(0.00)}  & \textbf{2.64(0.01)}  \\
\textbf{Wine}                                 & 0.93(0.01)          & 0.94(0.12)           & \textbf{0.89(0.05)}  & \textbf{0.86(0.02)}  \\
\textbf{Yacht}                                & 1.55(0.03)          & 1.18(0.21)           & \textbf{1.03(0.19)}  & \textbf{0.78(0.06)}  \\ \bottomrule
\end{tabular}}
\caption{RMSE and NLL of UCI regression benchmark datasets. The best score and the second score are highlighted in bolded. Standrad errors are reported in the parentheses.}
\label{tab1}
\end{table}

We first qualitatively evaluate the MT-ENet using a synthetic regression dataset. Fig~\ref{fig:fig3} represents our synthetic data and the model predictions. Experiments were carried out on an imbalanced regression dataset for two reasons: \textbf{1)} real-world regression tasks often involve an imbalanced target \cite{yang2021delving}; \textbf{2)} the ENet could fail on the imbalanced dataset. We observe that the predictions of the ENet for sparse samples (to the right sides of the green dotted lines in Fig~\ref{fig:fig3}) do not fit the data despite the well-calibrated uncertainty. This trend of the ENet (Fig \ref{fig:fig3} (c)) is in line with \textit{Theorem 1}, which stated that the MSE of the ENet is likely to be insufficient when the epistemic uncertainty is high. Conversely, the results obtained for other models, including the MT-ENet, are acceptable for the sparse sample regions. Training and experimental details are available in Appendix C.

\subsection{Performance evaluation on real-world benchmarks}

We evaluate the performance of the MT-ENet in comparison with strong baselines through the UCI regression benchmark datasets. The experimental settings and model architecture used in this study are identical to those used by \cite{hernandez2015probabilistic}. We report the average root mean squared error (RMSE) and the negative log-likelihood (NLL) of the models in Table \ref{tab1}. The MT-ENet generally provides the best or comparable RMSE and NLL performances. This is the evidence that the MT-ENet and the Lipschitz modified MSE loss improve the predictive accuracy of the ENet without disturbing the original NLL loss function. Details of the experiment are given in Appendix C.

\subsection{Drug-target affinity regression}

\begin{figure}[t]
    \frenchspacing
    \centering
    \includegraphics*[width=0.47\textwidth, clip=true]{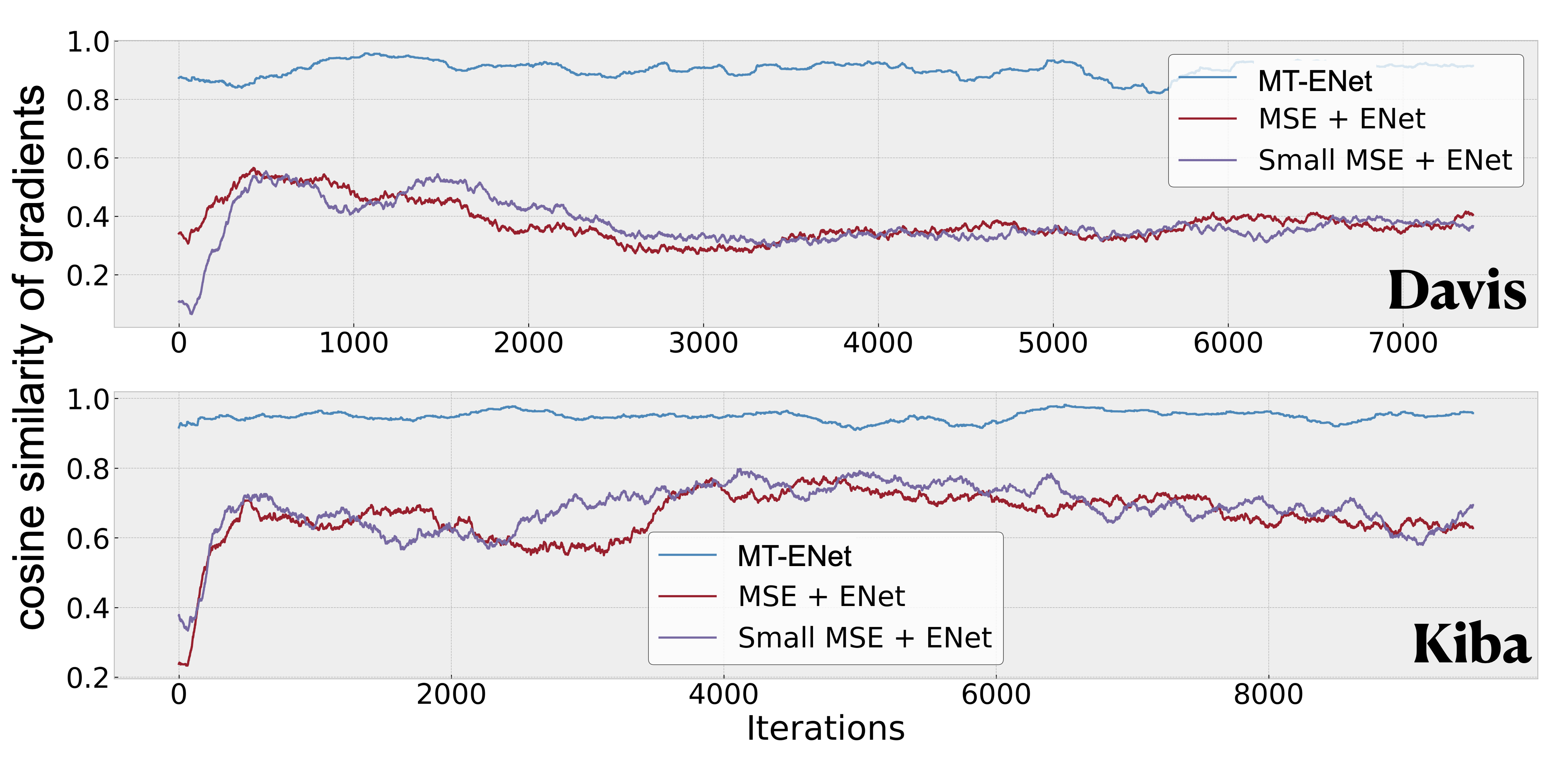}
    \caption{Cosine similarity trends between gradients from the additional loss and NLL. (Blue) the MT-ENet with the modified MSE; (Red) the ENet with the simple MSE; (Purple) the ENet with the weighted (0.1) MSE.}
    \label{fig:conflict}
\end{figure}
\frenchspacing
Next, we evaluate the MT-ENet on a high dimensional complex drug-target affinity (DTA) regression, which is an essential task for early-stage drug discovery \cite{shin2019self}. Our experiments use two well-known benchmark datasets in the DTA literature: Davis \cite{davis2011comprehensive} and Kiba \cite{tang2014making}. The Davis and Kiba datasets consist of protein sequence data and simplified molecular line-entry system (SMILES) sequences. The target value of the data is the drug-target affinity, which is a single real value.

The model architecture of the baselines and the MT-ENet, referred to as DeepDTA \cite{ozturk2018deepdta}, is the same except for the number of outputs: four outputs represent ($\gamma,\nu,\alpha,\beta$) for the ENet and MT-ENet; and a single output represents the target value for the MC-Dropout. The inputs of the models are the SMILES and protein sequences. The DeepDTA architecture consists of one-dimensional convolution layers for embedding the input sequences and fully connected layers to predict the target values using the embedded features. The details of training and the model architectures are available in Appendix C.

\subsubsection{Metrics} Our evaluation metrics are the MSE, NLL, ECE (Expected Calibration Error) and CI (Concordance Index). The MSE and NLL are typical losses in the optimizer. The role of the CI is to quantify the quality of rankings and the predictive accuracy of the model \cite{steck2008ranking, yu2011learning}. The CI is defined as:
$$
CI = \frac{1}{N} \sum_{y_i > y_j} h(\hat{y_i} > \hat{y_j}), \quad 
h(x) =\begin{cases}\small
1,&x>0\\
0.5,& x = 0\\
0,& \text{else}\\
\end{cases}
$$
where $N$ is the total number of data pairs; $y_i$ is a ground truth, and $\hat{y_i}$ is a predicted value. The ECE indicates how accurately the estimated uncertainty reflects real accuracy \cite{guo2017calibration, kuleshov2018accurate}. The ECE can be calculated as: 
$$
ECE = \frac{1}{N} \sum_{i=1}^N |acc(P_i) - P_i|
$$
Where $acc(P_i)$ is accuracy using $P_i$ confidence interval (e.g. $P_i=0.95$) and $N$ is the number of intervals. We use $P_{[0:N]}=\{0.01, 0.02 \dots, 1.00\}, N=100$.

\subsubsection{Alleviated gradient conflict by the Modified MSE}
We first demonstrate that the Lipschitz modified MSE can alleviate the gradient conflict by measuring cosine similarities of the gradient vectors. Note that, in this part of study, we use the ENet with the simple MSE as the additional loss (MSE ENet) as the baseline to determine whether our proposed Lipschitz loss can alleviate the conflict.

The moving average of the cosine similarity between gradients is shown in Fig~\ref{fig:conflict}, for 500 iterations averaged. The trends of the MT-ENet indicate the high cosine similarities, which is evidence that the Lipschitz modified MSE successfully alleviates the gradient conflict. Also, the results show that simply adjusting the weight of the MSE loss (the purple lines, small MSE) fails to resolve the conflict problem. Since the Lipschitz MSE is designed to avoid the conflict when the NLL loss results in the high predictive uncertainty high (\textit{B} in Fig~\ref{fig:illust_nll}), this experimental result is in good agreement with our conjecture regarding the MT-ENet gradient conflict: In the event that the NLL increases the predictive uncertainty of the model, then the gradient conflict is highly likely.

\subsubsection{Performance evaluation} 
\begin{table}[t]
\centering
\resizebox{\columnwidth}{!}{
\begin{tabular}{lllll}
\hline
\toprule
     & \multicolumn{4}{c}{\textbf{Davis}}                                    \\
     & ENet          & MT ENet       & MSE ENet      & Dropout      \\ \hline
CI   & 0.856(0.02)  & \textbf{0.864(0.01)}  & 0.863(0.02)  & \textbf{0.884(0.00)} \\
MSE  & 0.275(0.00)  & 0.273(0.01)  & \textbf{0.266(0.01)}  & \textbf{0.248(0.01)} \\
NLL  & -2.344(0.42) & -\textbf{2.424(0.07)} & \textbf{-2.430(0.10)} & 0.633(0.02) \\
ECE  & 0.184(0.02)  & \textbf{0.156(0.03)}  & \textbf{0.179(0.02)}  & 0.217(0.01) \\
\hline
\toprule
     & \multicolumn{4}{c}{\textbf{Kiba}}                                     \\
     & ENet       & MT ENet       & MSE ENet      & Dropout      \\
\midrule
CI   & 0.885(0.00)  & \textbf{0.887(0.00)}  & \textbf{0.888(0.00)}  & 0.872(0.00) \\
MSE  & 0.190(0.00)  & 0.181(0.00)  & \textbf{0.176(0.00)}  & \textbf{0.178(0.00)} \\
NLL  & \textbf{-1.544(0.05)} & \textbf{-1.433(0.07)} & -1.276(0.04) & 0.465(0.01) \\
ECE  & \textbf{0.077(0.03)}  & \textbf{0.066(0.01)}  & 0.081(0.02)  & 0.162(0.01) \\
\hline
\toprule
{} & \multicolumn{4}{c}{\textbf{BindingDB}}                      \\
{} &       ENet &    MT ENet &        MSE ENet &       Dropout \\
\midrule
CI  &  0.822(0.00) &  0.824(0.00) &  \textbf{0.831(0.00)} &  \textbf{0.830(0.00)} \\
MSE &  0.704(0.01) &  0.694(0.00) &  \textbf{0.637(0.01)} &  \textbf{0.641(0.01)} \\
NLL &  \textbf{0.944(0.02)} &  \textbf{0.937(0.01)} &  0.972(0.01) &  1.90(0.05) \\
ECE &  \textbf{0.018(0.01)} &  \textbf{0.015(0.00)} &  0.027(0.01) &  0.134(0.01) \\
\bottomrule
\end{tabular}}
\caption{The performance evaluation results on the DTA benchmark datasets. \textit{MSE ENet} represents the ENet using the simple MSE as the additional loss. \textit{Dropout} represents the MC-Dropout. Standard deviations are reported in the parentheses.}
\label{tab2}
\end{table}

Table \ref{tab2} represents the performance evaluation results of the Kiba and Davis datasets. The MT-ENet and MSE ENet successfully improve the predictive accuracy metrics (CI, MSE) for both datasets. This result confirms that the additional MSE-based loss functions can improve the point-estimation capability of the ENet as described in Sec~\ref{sec:3}. Note also that the MT-ENet enhances accuracy without any significant degradation of the uncertainty estimation capability, which is indicated by the NLL and ECE. Notably, in the case of the calibration metric (ECE), the MT-ENet shows the best results among the existing models. Conversely, the NLL and ECE of the MSE-ENet on the Kiba dataset are degraded to some extent. This sacrifice of the uncertainty estimation ability of the MSE-ENet can likely be attributed to the gradient conflict with the NLL loss function.

\paragraph{Out-of-distribution testing on the BindingDB dataset}

\begin{figure}[t]
    \centering
    \includegraphics*[width=\columnwidth, clip=true]{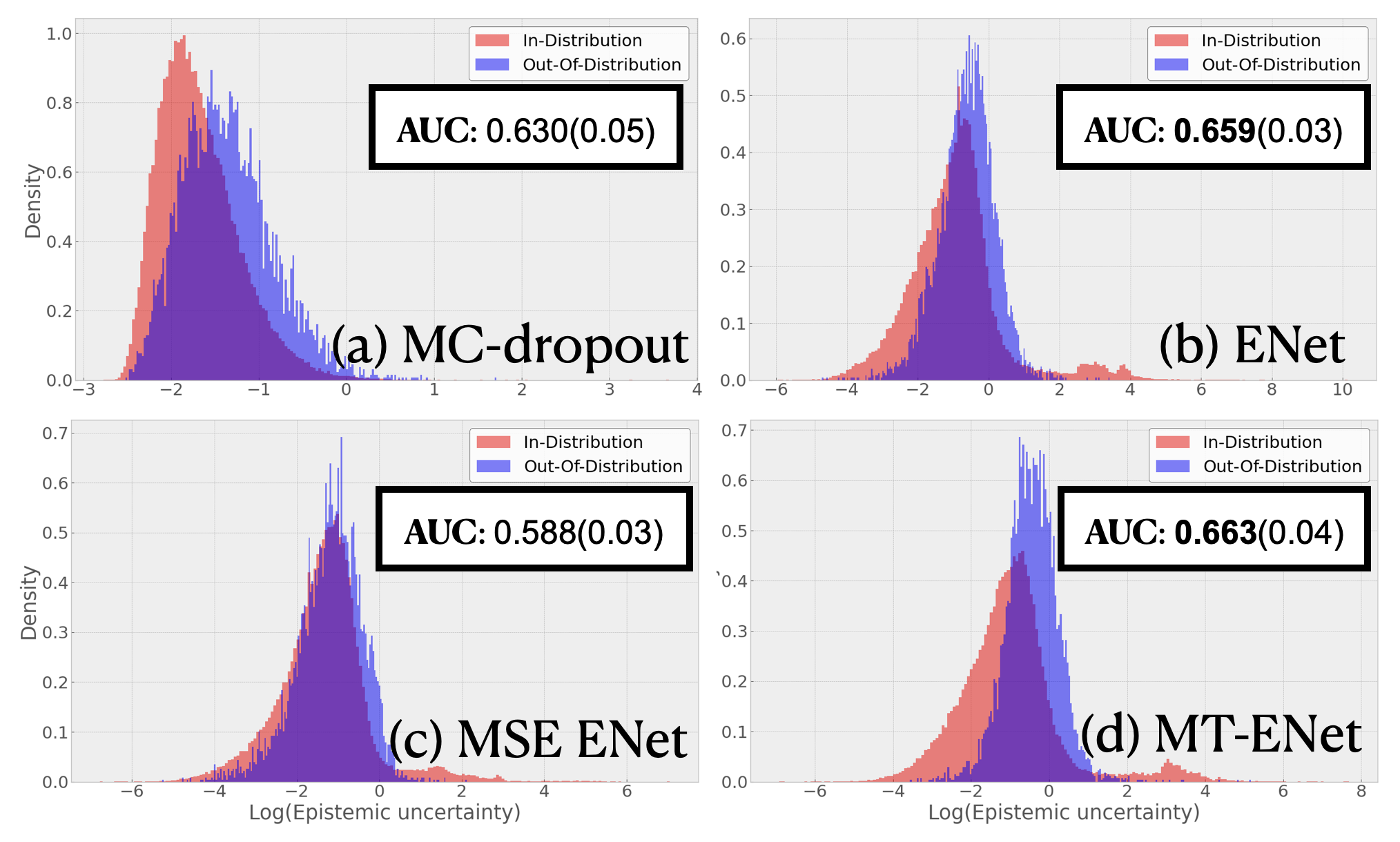}
    \caption{Density plots of log epistemic uncertainty. The AUC-ROC is also reported. The red distributions represent the ID data; The blue distributions represent the OOD data. \textit{MSE ENet} represents the ENet with the simple MSE as the additional loss.}
    \label{fig:ood}
\end{figure}

We examine the out-of-distribution (OOD) detection capability of the MT-ENet on the curated BindingDB dataset \cite{liu2007bindingdb}. The BindingDB dataset includes various drug-target (molecule-protein) sequence pairs. Because we excluded kinase protein samples---which are the target of the Kiba dataset---pairs of the kinase proteins from the Kiba dataset and detergent molecules from the PubChem \cite{kim2019pubchem} are considered as the OOD data. The test dataset of the BindingDB is considered as the in-distribution (ID) data.

For a total of three times, we randomly split the BindingDB dataset into the training (80\%), validation (10\%), and test (10\%) datasets. All examined models are trained three times with the training datasets. The details of the experiments and these datasets are available in Appendix B.

Fig~\ref{fig:ood} represents the distributions of log epistemic uncertainty and the AUC-ROC score of the examined models. The MT-ENet and ENet show high AUC-ROC scores, which implies excellent detection capability for OOD samples. In addition, Table~\ref{tab2} reports the performance for the test dataset of the BindingDB. Table~\ref{tab2} and Fig~\ref{fig:ood} show that the MT-ENet improves the ENet in all the metrics on the BindingDB dataset, including the calibration and OOD detection. In contrast, the AUC-ROC and ECE of the MSE-ENet considerably underperform in comparison to other methods, though the MSE-ENet shows the best MSE and CI on the BindingDB dataset. The exception is that the ECE and NLL of the MSE-ENet are better than that of the MC-dropout. This result is evidence that an additional loss function (the simple MSE) for the MSE ENet, which does not avoid the gradient conflict, can result in an overconfidence problem since it could disturb the uncertainty estimation of the NLL loss.

Furthermore, the density plots of the aleatoric uncertainty of the MSE-ENet, ENet, and MT-ENet are provided in Appendix D. The results demonstrate that the aleatoric uncertainties are at a similar level for the OOD and ID data. These results of the experiment are evidence that the MT-ENet is capable of disentangling the aleatoric and the epistemic uncertainties.

\section{Conclusion}
We have theoretically and empirically demonstrated that the MSE-based loss improves the point estimation capability of the ENet by resolving the gradient shrinkage problem. We proposed the MT-ENet which uses the Lipschitz modified MSE as the additional training objectives to ensure that the uncertainty estimation of the NLL is not disturbed. Our experiments show that the MT-ENet improves predictive accuracy not only maintaining but also sometimes enhance the uncertainty estimation of the ENet. Successful real-world deep learning systems must be \textit{accurate}, \textit{reliable}, and \textit{efficient}. We look forward to improving the evidential deep learning---that is, \textit{efficient} and \textit{reliable}---with better \textit{accuracy} through the MTL framework. 

\section*{Acknowledgments} This research was results of a study on the \textit{HPC Support} Project, supported by the \textit{Ministry of Science and ICT} and \textit{NIPA (National IT Industry Promotion Agency - Republic of Korea)}.

\bibliography{aaai22}

\newpage
\begin{center}
    \Huge \textbf{Appendix}
\end{center}
\appendix
\renewcommand\thefigure{A\arabic{figure}}
\renewcommand\theequation{A\arabic{equation}}
\frenchspacing
\maketitle

\section{Derivations and proofs}
In this section, we provide the proofs of the theorems and propositions in the main manuscript. Before we give the proofs, we rewrite the evidental neural network and its outputs and Definitions from the main manuscripts.

\paragraph{Outputs of the evidential neural network} Let $f$ be the ENet, $\theta$ be the trainable parameters of the ENet, and $X\in\mathbb{R}^d$ is the input data where $d$ is its dimension. The outputs of parameters of the ENet are $f(\theta,X)=\mathbf{m}=(\gamma,\nu,\alpha,\beta)$ where $\gamma\in\mathbb{R}, \nu>0, \alpha>1, \beta>0$. Specifically, the output of the ENet represents the NIG distribution:
$$
y \sim \mathcal{N}(\mu,\sigma),\quad \mu \sim \mathcal{N}(\gamma, \tfrac{\sigma^2}{\nu}),\quad  \sigma^2\sim \text{Gamma}^{-1}(\alpha,\beta) 
$$
The model prediction ($\mathbb{E}[\mu]$), aleatoric ($\mathbb{E}[\sigma^2]$), and epistemic ($Var[\mu]$) uncertainty of the ENet can be calculated by the following:
$$\mathbb{E}[\mu] = \gamma, \quad \mathbb{E}[\sigma^2] = \frac{\beta}{\alpha - 1}, \quad  Var[\mu] = \frac{\beta}{\nu(\alpha - 1)}.
$$

\paragraph{Definition 1.} \label{def2}
\textit{We define $\alpha$ and $\nu$, the outputs of the ENet, as the \textbf{pseudo observations}.}

\paragraph{Definition 2.} \label{def4} \textit{Consider a loss function $L:\mathbb{R}^n \rightarrow \mathbb{R}$ and a set of outputs of a model, $f_\theta(X) \in\mathbb{R}^n$, with parameters $\theta$. Let $\mathbf{\Omega}_\theta$ be the subset of the model outputs, $\mathbf{\Omega}_\theta \subseteq f_\theta(X)$, then $\mathbf{g}_{L,\mathbf{\Omega}} = \sum_{\omega_\theta \in \Omega_\theta} \tfrac{\partial L}{\partial\mathbf{\omega_\theta}} \nabla_\theta \mathbf{\omega_\theta}$ denotes the \textbf{gradient vector} with respect to the subset of the model outputs, $\mathbf{\Omega}_\theta$. We exclude the reliance of $\theta$ to simplify notations.}

\paragraph{Definition 3.} \textit{We define the \textbf{point estimation gradient} of the NLL as $\mathbf{g}_{L_{NLL},\gamma}$, since $\gamma$ performs the point estimation. The  \textbf{uncertainty estimation gradient} of the NLL is defined as $\mathbf{g}_{L_{NLL},\{\nu,\alpha,\beta\}}$, since $\{\nu,\alpha,\beta\}$ perform the uncertainty estimation. Finally, we define the \textbf{total gradient} of the NLL as $\mathbf{g}_{L_{NLL},\mathbf{m}}$.}

\subsection{Proof of theorem 1} We prove Theorem 1. Here, we first rewrite the NLL loss:
\begin{align*}
L_{NLL}(y,\textbf{m}) = \tfrac{1}{2}\log (\tfrac{\pi}{\nu}) - \alpha \log (2\beta(1+\nu))\\
+ (\alpha + \tfrac{1}{2})\log ((y - \gamma)^2\nu + 2\beta(1+\nu)) + \log(\tfrac{\Gamma(\alpha)}{\Gamma(\alpha + \tfrac{1}{2})})
\end{align*}

\paragraph{Theorem 1.} \label{them1} (Shrinking NLL gradient). \textit{Let $\nu>0, \gamma\in\mathbb{R}$ be the output of the ENet, and assume that $(y-\gamma)^2>0$, then for every real $\epsilon > 0$, there exists a real $\delta > 0 $ such that for every $\nu$, $0 < |\nu| < \delta$ implies $|\tfrac{\partial L_{NLL}}{\partial\gamma}| < \epsilon$. Therefore, $\lim_{\nu \rightarrow 0+}\tfrac{\partial}{\partial\gamma}L_{NLL}(y,\mathbf{m}) = 0$ $\Rightarrow$ $\lim_{\nu \rightarrow 0+}\|\mathbf{g}_{L_{NLL},\gamma}\| = 0$.}\\

(\textbf{Proof .})
Where $$\frac{\partial}{\partial\gamma}L_{NLL}(y,\mathbf{m}) = (2\alpha + 1)\frac{\lambda\nu}{\lambda^2 \nu + 2\beta(1+\nu)}$$, and $\lambda = \gamma - y$,
we show that if for every $\epsilon > 0$, there exists a $\delta > 0 $ such that, for all $\nu > 0$ if $0 < \nu < \delta$, then $|(2\alpha + 1)\frac{\lambda\nu}{\lambda^2\nu + 2\beta(1+\nu)}| < \epsilon$.
\begin{align*}
|\frac{\lambda\nu}{\lambda^2\nu + 2\beta(1+\nu)}| &< |(2\alpha + 1)\frac{\lambda\nu}{\lambda^2\nu + 2\beta(1+\nu)}| < \epsilon
\intertext{\hfill(since $\lambda^2\nu + 2\beta(1+\nu) > 0$)}
|\lambda|\nu=|\lambda\nu| &< \epsilon(\lambda^2\nu + 2\beta(1+\nu))\\
(|\lambda|-\epsilon\lambda^2-2\beta\epsilon)\nu &< 2\epsilon\beta
\intertext{If $(|\lambda|-\epsilon\lambda^2-2\beta\epsilon) < 0$, the proposition is always true regardless of $\delta$ since $\beta>0, 2\epsilon\beta>0$. Else:}
\nu &< \frac{2\epsilon\beta}{(|\lambda|-\epsilon\lambda^2-2\beta\epsilon)}
\end{align*}

Therefore, if we set $\delta \leq \frac{2\epsilon\beta}{(|\lambda|-\epsilon\lambda^2-2\beta\epsilon)}$, we say that
\begin{align*}
    \lim_{\nu \rightarrow 0+}\frac{\partial L_{NLL}}{\partial\gamma} &= \lim_{\nu \rightarrow 0+}(2\alpha + 1)\frac{\lambda\nu}{\lambda^2 \nu + 2\beta(1+\nu)} = 0\\
   \lim_{\nu \rightarrow 0+}\|\mathbf{g}_{L_{NLL},\gamma}\| &= \lim_{\nu \rightarrow 0+}\|\frac{\partial L_{NLL}}{\partial\gamma}[g_1,\dots,g_M]\|\\ &= \lim_{\nu \rightarrow 0+}\frac{\partial L_{NLL}}{\partial\gamma}\|[g_0,\dots,g_M]\| = 0 
\end{align*}
where $\theta$ is a set of trainable parameters of the ENet, $M=|\theta|$ and $g_{1:M}\in\mathbb{R}$.

\subsection{Proof of Proposition 1} 

We show that gradients of the NLL loss and MSE loss with respect to the model prediction ($\gamma$) are never conflict. Specifically, the cosine similarity between two gradients never  $< 1$. Then we provide the proof of Proposition 1:

\paragraph{Proposition 1.} (Non-conflicting point estimation). \textit{If the L2 norm of $\mathbf{g}_{L_{MSE},\gamma},\mathbf{g}_{L_{NLL},\gamma}$ are not zero, then the cosine similarity $s(\cdot)$ between $\mathbf{g}_{L_{MSE},\gamma}$, and $\mathbf{g}_{L_{NLL},\gamma}$ is always one. Hence, $\mathbf{g}_{L_{MSE},\gamma}$ is positively proportional to  $\mathbf{g}_{L_{NLL},\gamma}$}. 
\begin{equation}
\begin{gathered}
\label{eq9}
s(\mathbf{g}_{L_{MSE},\gamma},\mathbf{g}_{L_{NLL},\gamma}) = 1 \ \Rightarrow \quad \mathbf{g}_{L_{MSE},\gamma} = k\mathbf{g}_{L_{NLL},\gamma}\
\end{gathered}
\end{equation}
\textit{where $k > 0$}

(\textbf{Proof .}) We can express $\mathbf{g}_{L_{NLL},\gamma}$ and $\mathbf{g}_{L_{MSE},\gamma}$ as following:
$$
\mathbf{g}_{L_{MSE}, \gamma} =\tfrac{\partial L_{MSE}}{\partial\gamma} \nabla_\theta \gamma = \frac{\partial L_{MSE}}{\partial\gamma}[g_0,\dots,g_M]$$
$$
\mathbf{g}_{L_{NLL}, \gamma} = \tfrac{\partial L_{NLL}}{\partial\gamma} \nabla_\theta  \gamma = \frac{\partial L_{NLL}}{\partial\gamma}[g_0,\dots,g_M]
$$
where $g_{0:M} \in \mathbb{R}$. We can calculate the cosine similarity, $s(\cdot)$, between the two gradient vectors $\mathbf{g}_{L_{MSE},\gamma}$ and $\mathbf{g}_{L_{NLL},\gamma}$:

$$
s(\mathbf{g}_{L_{MSE},\gamma},\mathbf{g}_{L_{NLL},\gamma}) = \frac{\mathbf{g}_{L_{MSE},\gamma}\cdot \mathbf{g}_{L_{NLL},\gamma}}{\|\mathbf{g}_{L_{MSE},\gamma}\| \|\mathbf{g}_{L_{NLL},\gamma}\|}
$$
$$
= \frac{\tfrac{\partial L_{NLL}}{\partial\gamma}\tfrac{\partial L_{MSE}}{\partial\gamma} (g_0^2 + \dots + g_m^2)}{|\tfrac{\partial L_{NLL}}{\partial\gamma}| |\tfrac{\partial L_{MSE}}{\partial\gamma}| \sqrt{g_0^2 + \dots + g_m^2} \sqrt{g_0^2 + \dots + g_m^2}}$$
$$ = \frac{\tfrac{\partial L_{NLL}}{\partial\gamma}\tfrac{\partial L_{MSE}}{\partial\gamma}}{|\tfrac{\partial L_{NLL}}{\partial\gamma}| |\tfrac{\partial L_{MSE}}{\partial\gamma}|}
$$
The signs of the $\tfrac{\partial L_{NLL}}{\partial\gamma}, \tfrac{\partial L_{MSE}}{\partial\gamma}$ are always identical:
\begin{align*}
    \tfrac{\partial}{\partial\gamma}L_{NLL}(y,\mathbf{m}) &= \tfrac{\partial}{\partial \gamma} - \log t(y; \gamma, \frac{\beta(1+\nu)}{\nu \alpha}, 2\alpha) \\ &= (2\alpha + 1)\frac{\lambda\nu}{\lambda^2 \nu + 2\beta(1+\nu)}\\
    \tfrac{\partial}{\partial\gamma}L_{MSE}(y,\gamma) &= \tfrac{\partial}{\partial\gamma} (y-\gamma)^2 = 2\lambda
\end{align*}
$$\begin{cases} 
\text{If. } \lambda \geq 0 & \tfrac{\partial L_{NLL}}{\partial\gamma} \geq 0 , \tfrac{\partial L_{MSE}}{\partial\gamma} \geq 0
\\
\text{Else if. } \lambda < 0 & \tfrac{\partial L_{NLL}}{\partial\gamma} < 0 , \tfrac{\partial L_{MSE}}{\partial\gamma} < 0
\end{cases}$$
Therefore, we conclude: $s(\mathbf{g}_{L_{MSE},\gamma},\mathbf{g}_{L_{NLL},\gamma}) = 1$ and $ \mathbf{g}_{L_{MSE},\gamma} = k\mathbf{g}_{L_{NLL},\gamma} \quad k > 0$.

\subsection{Proof of Corollary 1} We prove Corollary 1, using Proposition1 and the following property of the gradient vector ,$\textbf{g}$:
\begin{itemize}
    \item Let $\mathbf{\Omega}$ be the set of outputs of a model and $L$ be a loss function. If there are two subsets $\mathbf{\hat{\Omega}}_1,\mathbf{\hat{\Omega}}_2$ of $\mathbf{\Omega}$, and $\mathbf{\hat{\Omega}}_1 \cap \mathbf{\hat{\Omega}}_2=\phi$, then $\mathbf{g}_{L,\mathbf{\hat{\Omega}_1\cup\hat{\Omega}_2}}=\mathbf{g}_{L,\mathbf{\hat{\Omega}}_1}+\mathbf{g}_{L,\mathbf{\hat{\Omega}}_1}$.
\end{itemize}

\paragraph{Corollary 1.} (Source of gradient conflict). \textit{If the \textbf{total gradients} of the NLL and the gradient of MSE are not in the same direction, then the \textbf{uncertainty estimation gradients} of the NLL  and the gradients of the MSE are not in the same direction:}
$s(\mathbf{g}_{L_{MSE},\mathbf{m}}, \mathbf{g}_{L_{NLL},\mathbf{\mathbf{m}}})<1 \ \Rightarrow \ s(\mathbf{g}_{L_{MSE},\mathbf{m}}, \mathbf{g}_{L_{NLL},\mathbf{\{\nu,\alpha,\beta\}}}) < 1$

(\textbf{Proof .}) We consider the contrapositive of the Corollary 1:

\paragraph{(Contrapositive of Corollary 1.)} \textit{If the gradients of the NLL loss from ( $\nu,\alpha,\beta$) and the gradients of the MSE loss are not conflict,  $s(\mathbf{g}_{L_{MSE},\mathbf{m}}, \mathbf{g}_{L_{NLL},\mathbf{\{\nu,\alpha,\beta\}}})=1$, then the gradients of two losses never conflict:  $s(\mathbf{g}_{L_{MSE},\mathbf{m}}, \mathbf{g}_{L_{NLL},\mathbf{\mathbf{m}}})=1$.}

On account of an assumption:
$s(\mathbf{g}_{L_{MSE}, \mathbf{m}}, \mathbf{g}_{L_{NLL} \mathbf{\{\nu,\alpha\,\beta\}}})=1$, we have $\mathbf{g}_{L_{MSE}, \mathbf{m}} = k'\mathbf{g}_{L_{NLL} \mathbf{\{\nu,\alpha\,\beta\}}}$ where $k' > 0$. From \textit{Theroem 2}, we get $\mathbf{g}_{L_{MSE},\gamma} = k\mathbf{g}_{L_{NLL},\gamma}$ where $ k > 0$. Since $\mathbf{g}_{L_{MSE},\gamma} = \mathbf{g}_{L_{MSE},\mathbf{m}}$, we have: 
\begin{align*}
s(&\mathbf{g}_{L_{MSE}, \mathbf{m}},\mathbf{g}_{L_{NLL} \mathbf{m}}) = \frac{\mathbf{g}_{L_{MSE}, \mathbf{m}}\cdot \mathbf{g}_{L_{NLL} \mathbf{m}}}{\|\mathbf{g}_{L_{MSE}, \mathbf{m}}\|\|\mathbf{g}_{L_{NLL} \mathbf{m}}\|}\\
=&\frac{\mathbf{g}_{L_{MSE}, \mathbf{m}}\cdot (\mathbf{g}_{L_{NLL} \gamma} + \mathbf{g}_{L_{NLL} \mathbf{\{\nu,\alpha\,\beta\}}})} {\|\mathbf{g}_{L_{MSE}, \mathbf{m}}\|\|\mathbf{g}_{L_{NLL} \gamma} + \mathbf{g}_{L_{NLL} \mathbf{\{\nu,\alpha\,\beta\}}}\|}\\
=&\frac{\mathbf{g}_{L_{MSE}, \mathbf{m}}\cdot (\tfrac{(1+k')k}{k'})\mathbf{g}_{L_{NLL} \gamma}} {\|\mathbf{g}_{L_{MSE}, \mathbf{m}}\|\|\tfrac{(1+k')k}{k'}\mathbf{g}_{L_{NLL} \gamma}\|}\\
=&s(\mathbf{g}_{L_{MSE},\gamma},\mathbf{g}_{L_{NLL},\gamma}) = 1 \qquad \text{By Proposition 1}
\end{align*}

Since the contrapositive of Corollary 1 is true, we conclude that Corollary 1 is true.

\subsection{Proof of Proposition 2} In this section, we proof Proposition 2. Let $\lambda^2 = (\gamma - y)^2$, which represents the squared error value. Then the gradient of the pseudo observations, $\alpha, \nu$, of the NIG distribution w.r.t the NLL loss $L_{NLL}$ is positive when the $\lambda^2$ is larger then $U_\alpha$ or $U_\nu$:

\paragraph{Proposition 2.} \label{prop2} \textit{Let $\lambda^2\eqdef(y-\gamma)^2$, which is the squared error value of the ENet. then if $\lambda^2$ is larger than certain thresholds, $U_\nu$ and $U_\alpha$, then the derivative signs of the $L_{NLL}$ w.r.t $\nu$ and $\alpha$ are positive.}
\begin{equation}
\label{eq9}
\begin{gathered}
\lambda^2 >  U_{\alpha} \Rightarrow \frac{\partial}{\partial\alpha} L_{NLL} > 0, \quad
\lambda^2 > U_{\nu} \Rightarrow  \frac{\partial}{\partial\nu} L_{NLL} > 0\\
U_{\nu} = \tfrac{\beta(\nu+1)}{\alpha\nu},\ U_{\alpha} = \tfrac{2\beta(1+\nu)}{\nu}[\exp(\Psi(\alpha + \tfrac{1}{2}) - \Psi(\alpha)) - 1]
\\
\end{gathered}
\end{equation}
\textit{where $\Psi(\cdot)$ is the digamma function.\\}

(\textbf{Proof .})
\begin{gather*}
\frac{\partial L_{NLL}}{\partial\alpha} = \log(1 + \frac{\lambda^2\nu}{2\beta(\nu+1)}) + \Psi(\alpha) - \Psi(\alpha + \tfrac{1}{2}) > 0 \\
\begin{align*}
\log(\frac{\lambda^2\nu}{2\beta(\nu+1)} + 1) &> \Psi(\alpha + \tfrac{1}{2}) - \Psi(\alpha)\\
\frac{\lambda^2\nu}{2\beta(\nu+1)} + 1 &> \exp(\Psi(\alpha + \tfrac{1}{2}) - \Psi(\alpha))\\ 
\frac{\lambda^2\nu}{2\beta(\nu+1)} &> \exp(\Psi(\alpha + \tfrac{1}{2}) - \Psi(\alpha)) - 1 
\end{align*}\\
\lambda^2 > [\exp(\Psi(\alpha + \tfrac{1}{2}) - \Psi(\alpha)) - 1]2\beta(1+\nu)\nu^{-1} \eqdef U_\alpha 
\end{gather*}

and,

\begin{gather*}
\frac{\partial L_{NLL}}{\partial\nu} = - \tfrac{1}{2\nu} - \tfrac{\alpha}{\nu+1} + (\alpha + \tfrac{1}{2})\frac{\lambda^2 + 2\beta}{\lambda^2\nu + 2\beta(1+\nu)} > 0 \\
\begin{align*}
\tfrac{1}{2\nu} + \tfrac{\alpha}{1+\nu} < (\alpha +& \tfrac{1}{2})\frac{\lambda^2 + 2\beta}{\lambda^2\nu + 2\beta(1+\nu)}\\
(\tfrac{1}{2\nu} + \tfrac{\alpha}{1+\nu})(\lambda^2\nu + 2\beta(1+\nu)) &< (\alpha + \tfrac{1}{2})(\lambda^2 + 2\beta)\\
\alpha(\tfrac{\nu}{1+\nu} - 1)\lambda^2 &< 2\beta(\alpha + \tfrac{1}{2} - \tfrac{1+\nu}{2\nu} - \alpha)\\
\alpha(\tfrac{\nu}{1+\nu} - 1)\lambda^2 &< 2\beta(1-\tfrac{1+\nu}{\nu})
\end{align*}\\
\lambda^2 > \frac{\beta(1-\tfrac{1+\nu}{\nu})}{\alpha(\tfrac{\nu}{1+\nu} - 1)} = \frac{\beta(\nu+1)}{\alpha\nu} \eqdef U_{\nu}
\end{gather*}


\section{Evidential regularization}
The evidential regularization, $L_R$, of the ENet was originally proposed in  \cite{amini2020deep}. 
\begin{equation}
    L_R(y,\mathbf{m}) = |y - \gamma|(2\nu + \alpha)
    \label{eq:appendix_LR}
\end{equation}
The role of the $L_R$ is to regularize the NLL loss, $L_{NLL}$, by increasing the predictive uncertainty of incorrect predictions. In particular, if the difference between the model prediction, $\gamma$, and the true value, $y$, is large, it tends to decrease the pseudo observations, ($\alpha,\nu$):
\begin{equation}
\frac{\partial L_R}{\partial\nu} = 2|y-\gamma|, \quad \frac{\partial L_R}{\partial\alpha} = |y-\gamma|
\label{eq:appendix_av}
\end{equation}
As we mentioned in Sec 2, the decrease in the pseudo observations leads to increase in the predictive uncertainty because they are inversely proportional: $\mathbb{E}[\sigma^2] \propto \tfrac{1}{\alpha}$ and $Var[\mu] \propto \tfrac{1}{\alpha \nu}$. Therefore, the evidential regularizer increases the uncertainty of the incorrect predictions, which has large $|y-\gamma|$.

Moreover, $L_R$ also depends on $\gamma$, which represents the model prediction $\mathbb{E}[\mu]=\gamma$, as we can noticed from eq~\ref{eq:appendix_LR}. The derivative of $L_R$ with respect to $\gamma$ is following:
$$
\frac{\partial L_R}{\partial\gamma} = 
\begin{cases}
(2\nu + \alpha), & \text{if} \quad y - \gamma < 0\\
-(2\nu + \alpha), & \text{else if} \quad y - \gamma \geq 0\\
\end{cases}$$
The derivatives of $L_R$ with respect to $\gamma$ do not depend on the difference between the model prediction and the true value, $y-\gamma$, but depend on the pseudo observations, $\nu,\alpha$. However, since $\nu$ and $\alpha$ could be decreased due to eq~\ref{eq:appendix_av} for the incorrect predictions, $\tfrac{\partial L_R}{\partial \gamma}$ is also being small. It implies that the gradient of the $L_R$, $\mathbf{g}_{L_R,\gamma}$, could shrink---similar to Theorem 1---. Thus, the $L_R$ could not effectively train the ENet to predict $y$. It suggests that the $L_R$ cannot play a role of the MSE-based loss functions which is effectively training the model prediction of the ENet although the $L_R$ has $\gamma$ in its equation.

\begin{figure}
    \includegraphics[width=\columnwidth]{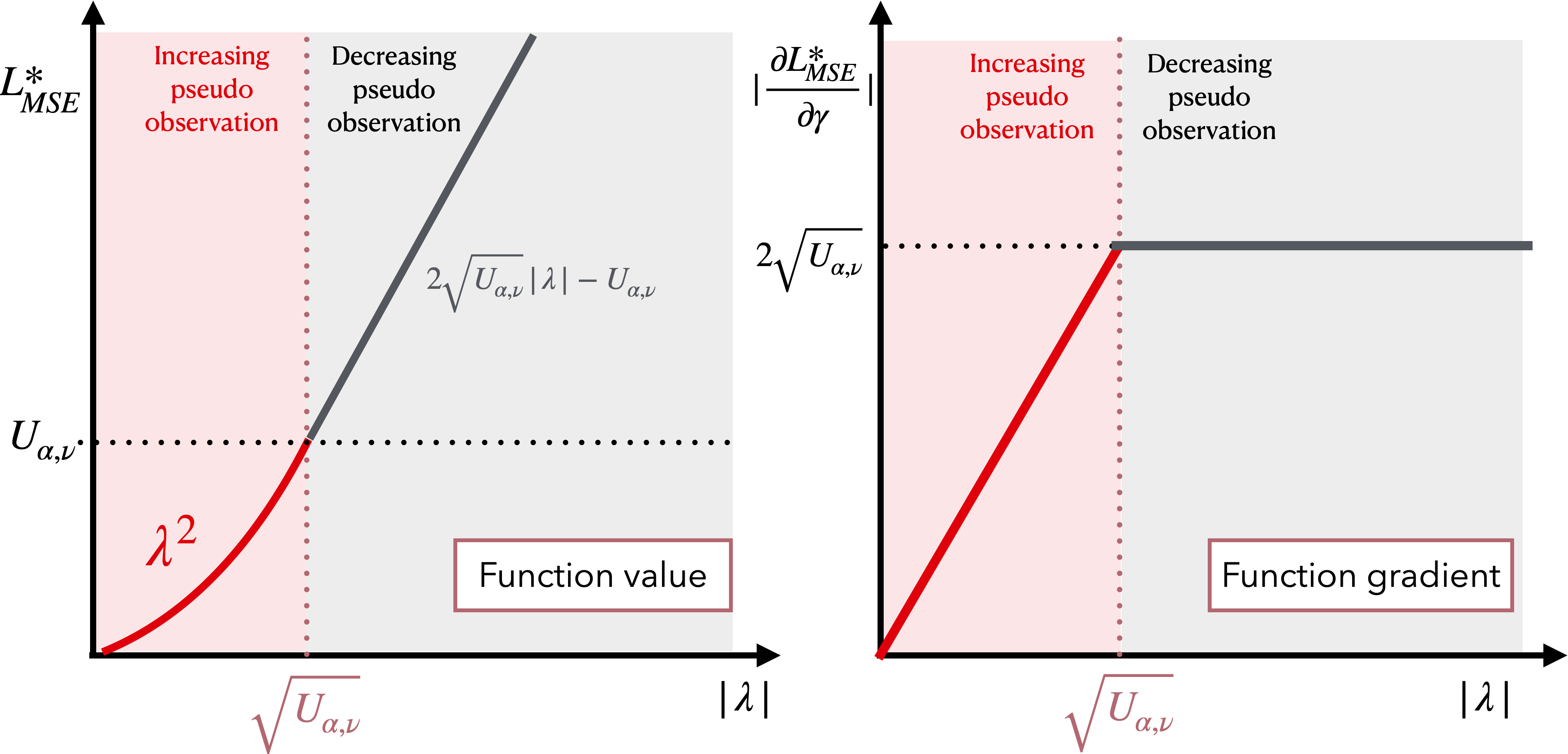}
    \caption{A scheme of the Lipschitz modifed MSE loss. $U_{\alpha,\nu}$ is the threshold value of the Sec 5.1 and $\lambda=\gamma-y$. The Lipschitz modified loss function $L^*_{MSE}$ restricts the maximum absolute value of the derivatives only when the MSE error is larger than the $U_{\nu,\alpha}$.}
    \label{fig:figs1}
\end{figure}

\section{Details of experiments}

\subsection{Synthetic dataset experiment}

We generate the synthetic dataset from the function: $y = sin(4x)^3 + x^2/10 + 2\epsilon + 0.1u; \ \ u \sim \text{Unif}
(x);\ \  \epsilon \sim \mathcal{N}(0,0.05 |7-|x||)$. We generate 2000 numbers of 1-D samples ranged [-3, 10]. Our generated samples consist of 1950 numbers of uniformly generated samples in the range [-3, 6] and 50 uniformly generated samples in the range [6,10]. The numbers of training samples are 900 and 20 for data-rich (The left blue shaded area in Fig 1) and for data-sparse regions (The right blue shaded area in Fig 1), respectively.

For the model architectures, we use three hidden layers for all examined models. Each hidden layer has 100 neurons and uses the tanh activation function. For the ENet and MT-ENet, the outputs are 4 times larger to represent the evidential parameters $\boldsymbol{m}=\{\gamma,\nu,\alpha,\beta\}$. For the hyperparameters, we use 0.01 as the learning rate of the Adam optimizer; $10^{-3}$ as the weight decay rate; $10^{-2}$ as the evidential regularizer factor; 128 as the minibatch size. Note that we exclude the weight decay for the MC-Dropout. The MC-dropout uses 0.2 as the dropout rate. We sample the MC-dropout outputs 5 times. The Gaussian process model uses the Matern5/2 kernel and 0.05 as its smoothness parameter. The Ensemble uses 5 differently initialized models.

\subsection{UCI regression benchmark experiment}
We evaluate the NLL and the MSE for the MC-dropout \cite{gal2016dropout}, ENet, Deep-Ensemble \cite{lakshminarayanan2017simple} and the MT-ENet using several UCI-regression benchmarks. We use the experimental setting of  \cite{hernandez2015probabilistic}. The original studies of the MC-dropout and Deep-Ensemble also reported their results using this setting \cite{gal2016dropout, lakshminarayanan2017simple}. 

The experiments of this paper and \citeauthor{gal2016dropout, hernandez2015probabilistic, lakshminarayanan2017simple, amini2020deep} follow steps: (1) hyperparameter optimization 30 times using Bayesian optimization (BO) for validation data; (2) training the model with the optimized hyperparameters; (3) evaluating the model with the test data and trained data. Each process consumes 40 epochs. The considered hyperparameters for the MT-ENet are learning-rate, and evidential regularization factor \cite{amini2020deep}. Furthermore, we normalize the input features and the target value using z-score normalization. Note that the normalization of the target value is excluded during the testing time. We use batch size 32 for every experiment. Since we followed the identical experiments, we reported the scores of the MC-dropout, Deep Ensemble, and the ENet, by using reported scores in their papers,
To summarize, the experimental steps are following:
\begin{itemize}
    \item Split the dataset the training dataset and the test dataset with 0.9 and 0.1 respectively.
    \item Split the training dataset again into 0.8 and 0.2, and optimize the hyperparameters w.r.t the validation dataset (0.2) with the trained model with the training dataset (0.8).
    \item Evaluate the test dataset with the model which was optimized hyperparameters w.r.t the validation dataset.
\end{itemize}
We repeat this procedure 20 times and report the MSE and the NLL, which is identical setting with \cite{gal2016dropout}, \cite{hernandez2015probabilistic}, \cite{lakshminarayanan2017simple} and \cite{amini2020deep}.

\subsection{Drug-target affinity prediction task}

\begin{figure}[t]
    \centering
    \includegraphics[width=\columnwidth]{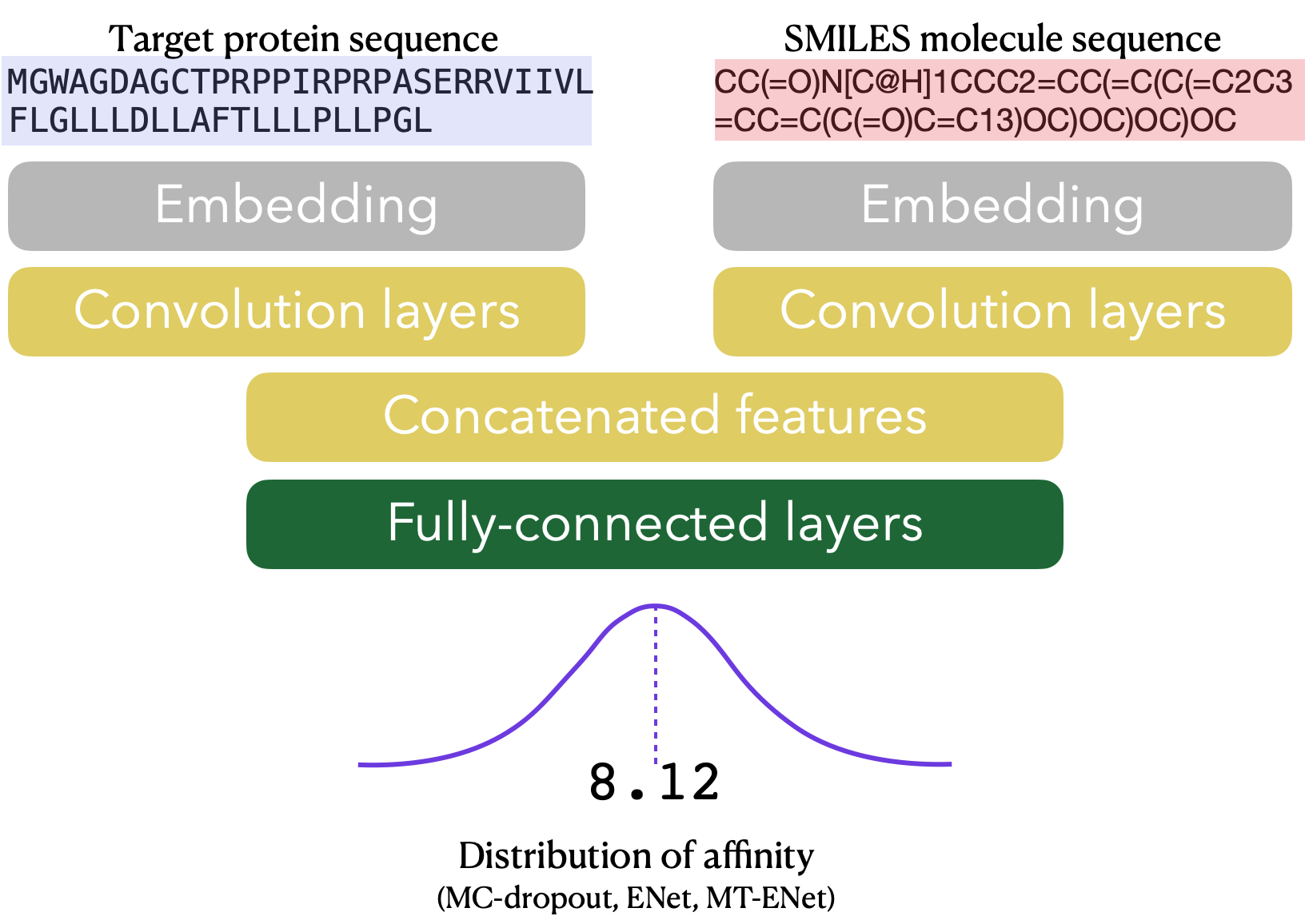}
    \caption{\small A scheme of the drug-target affinity prediction task and the DeepDTA architecture. The input data is a pair of a target protein sequence and a molecule sequence, and the output data is an affinity value of the pair. In this paper, the output of the DeepDTA is the likelihood distribution of the affinity, not a single real value of the affinity.}
    \label{fig:deepdta}
\end{figure}

For all DTA experiments, we train the model, and the best model is saved using the loss value with the validation dataset. The reported metrics in Table 2 use the trained model and the test dataset.

\paragraph{Datasets} We use two different benchmark datasets of Drug-Target-Affinity prediction tasks: Davis and Kiba. The benchmarks dataset consists of the kinase proteins and their small molecule inhibitors. Both only use the sequence information of proteins (target) and small molecules (drug), one is the protein residues sequences, and the other is the SMILES (Simplified Molecular-Input Line-Entry System) representations of the molecules. Both benchmark datasets are widely evaluated for several sequence-based drug-target interaction studies \cite{ozturk2018deepdta, shin2019self, pahikkala2015toward, he2017simboost, agyemang2020multi, zhao2020gansdta, abdel2020deeph, zeng2021deep}.

\begin{itemize}
    \item \textbf{Davis} dataset consists of clinically relevant kinase inhibitor ligands and their dissociation constant ($K_d$) values(Affinity values). Since the scale of the target affinity values is too varying, the dataset uses the negative log values ($-\tfrac{\log K_d}{1e^6}$) to resolve numerical issues. The Davis dataset has 68 compounds and 442 proteins, a total of 30,056 interactions (affinities).
     \item \textbf{Kiba} dataset also includes kinase proteins and small molecules. The target value of the Kiba data represents its affinity value. The Kiba datasets use their own affinity metric, called the Kiba score. The Kiba dataset 2,111 compounds and 229 proteins, a total of 118,254 interactions (affinities).
\end{itemize}

\paragraph{Remark (Target value of the Davis).} \textit{Davis dataset uses the negative log affinity value, $-\tfrac{\log K_d}{1e^6}$, as the target value. However, the original study of the Davis dataset uses 5 as the target value if the value is lower than 5, which could lead to noisy labels. Since the noisy labels are the cause of a high calibration error---according to \cite{nixon2019measuring}--- the high ECE in Table 2 supports that the Davis has the noisy labels. And the noisy labels of the Davis dataset can explain the outstanding performance of the MC-Dropout in our experiment (Table 2) since the MC-Dropout is robust against the noisy labels \cite{goel2021robustness}.}

\paragraph{Model} All examined models have the DeepDTA architecture as the backbone. The DeepDTA architecture has embedding layers to encode one-hot represented protein sequences and SMILES sequences, as shown in Fig~\ref{fig:deepdta}. Next, the model has the 1-d convolutional layers and global max-pooling layers to embed the SMILES sequences (molecule-sequence) and the protein sequences for each. Finally, three fully connected layers use the concatenated latent features of the proteins and the drugs to predict the affinity value. The DeepDTA uses ReLU as its activation function. We use 0.1 as the dropout probability, which is an identical value of the DeepDTA study. Note that only the fully connected layers use the dropout. The outputs of the ENet and the MT-ENet represents the evidential parameters of the NIG distribution, and the number of the output is four ($\gamma, \nu, \alpha, \beta$). In contrast, the MC-dropout uses a single output to represent the affinity value.

\paragraph{Hyperparameters and training details} We use 1E-4 as the evidential regularization factor and 1E-4 as the L2 regularization factor; We use 256 as the batch size; 200 as the number of epochs.

\subsection{Out-of-distribution experiment}
\begin{figure}[t]
    \centering
    \includegraphics[width=\columnwidth]{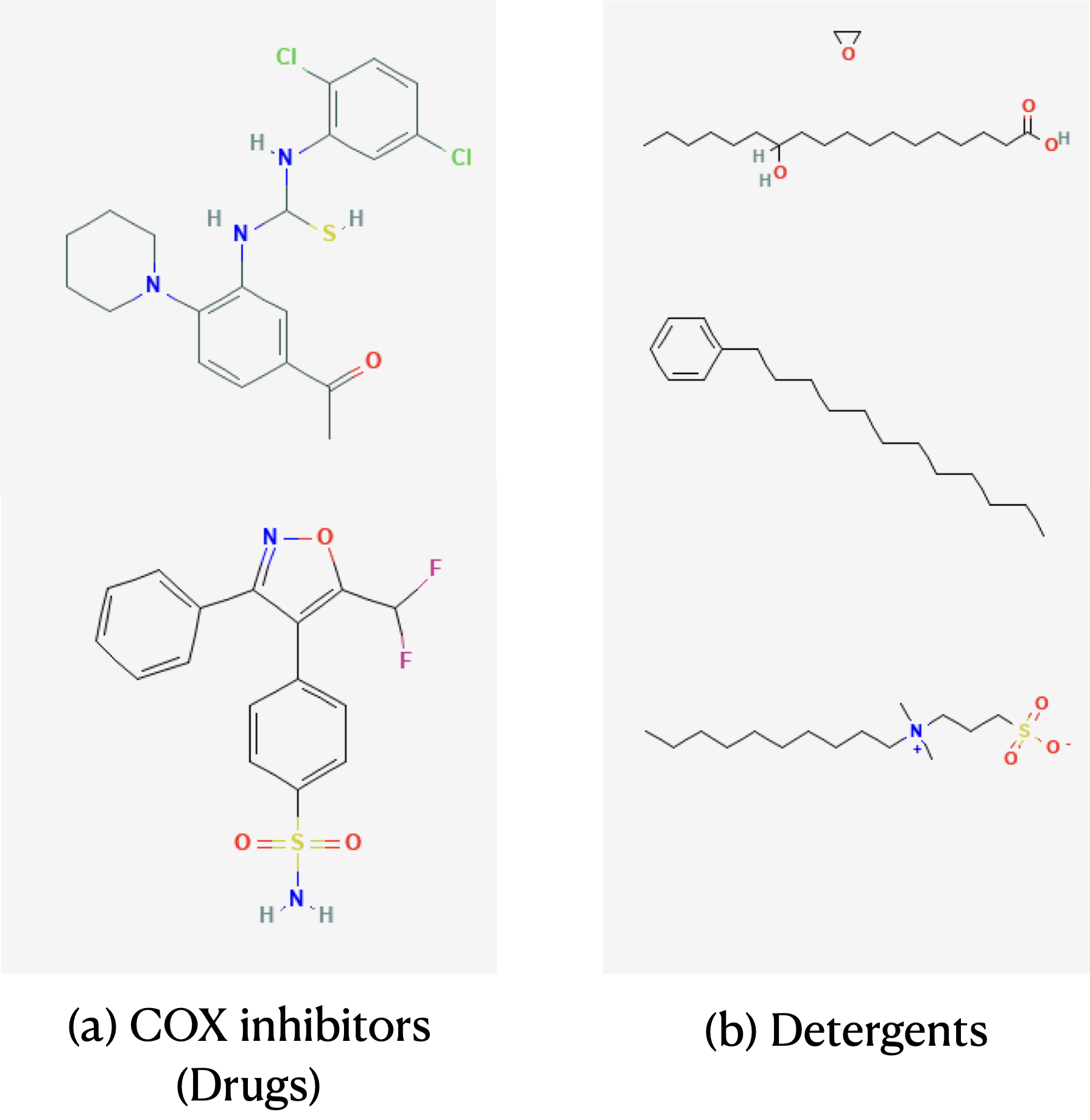}
    \caption{Molecule structure examples of the small molecule drugs (a) and the detergent molecules (b). The BindingDB dataset includes the small molecule drugs and their target protein pairs, such as (a).}
    \label{fig:molex}
\end{figure}

\paragraph{Datasets} We use BindingDB \cite{liu2007bindingdb} dataset. Like the Kiba and the Davis dataset, the BindingDB dataset consists of protein sequences and SMILES molecule sequences. Among them, we extracted the molecule-target pairs having $IC50$ values among different affinity metrics (Metric of the target value such as $KI, Kd, IC50$). For numerical stability, $IC50$ values are converted to $-\log\tfrac{IC50}{1e^6}$, similar to the Davis dataset. Importantly,  to consider the kinase samples as the out-of-distribution (OOD) dataset, we exclude the kinase protein samples during the curation. Since the Kiba dataset only consists of the kinase proteins, we use the protein sequences of the Kiba dataset as the OOD protein sequences. Additionally, the detergent molecule sequences from PubChem are used as the OOD SMILES molecule sequences \cite{kim2019pubchem}. Fig~\ref{fig:molex} represents the example structures of detergents molecules and drug molecules (COX inhibitors).  The input of the OOD samples is the pair of the kinase protein sequences and the detergent molecule sequences. For the in-distribution (ID) dataset, the test set of the BindingDB dataset is used.

\paragraph{Models} We use identical model architectures with the Kiba and Davis experiments.

\paragraph{Hyperparameters and training details} We randomly split the BindingDB dataset into 0.8, 0.1, and 0.1 as training, validation, and test dataset three times. The models learn the training dataset, and the best model is saved using based on the validation loss. We train the models with the following hyperparameters: 0.001 as the learning rate; 0.001 as the evidential regularization factor; 0.1 as the dropout probability for the MC-dropout; 100 as the number of epochs; 256 as the batch size. The model architectures are identical to the Kiba and the Davis experiments.

\section{Aleatoric uncertainty on the BindingDB dataset}
\begin{figure}[h]
    \centering
    \includegraphics[width=0.9\columnwidth]{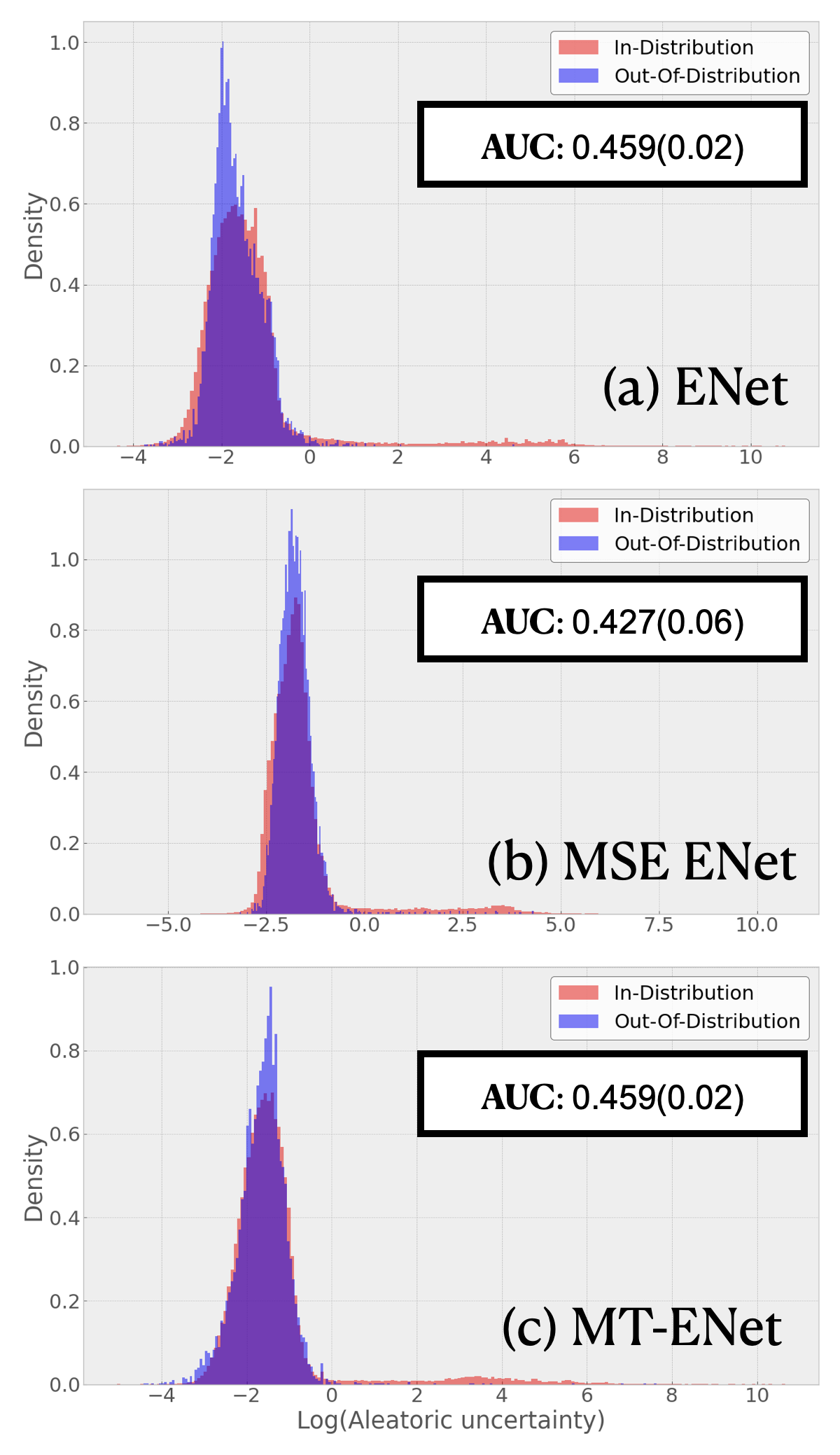}
    \caption{The distributions of the log aleatoric uncertainty of the ENet, MSE-ENet, and MT-ENet. According to the definition of the aleatoric uncertainty and its empirical evidence \cite{gal2016uncertainty, kendalluncertainties}, the aleatoric uncertainty does not increase for the OOD samples. All examined models show these trends.}
    \label{fig:alea}
\end{figure}
In this section, we show that the ENet and MT-ENet are capable of that distinguishing the aleatoric uncertainty and the epistemic uncertainty. The original study of the ENet \cite{amini2020deep} designed to have the capability of the ENet distinguishing the epistemic uncertainty and aleatoric uncertainty. According to \cite{gal2016uncertainty}, the aleatoric uncertainty is the irreducible uncertainty in data observation. Thus, the aleatoric uncertainty should not increase for the out-of-distribution (OOD) samples.

Fig~\ref{fig:alea} represents the predictive entropy of the aleatoric uncertainty of the ENet, MT-ENet, and MSE-ENet. We use the same models and datasets of the experiment of sec 3.3 \textit{(Out-of-distribution testing on the BindingDB dataset)}. Unlike Fig 6, the aleatoric uncertainty of the OOD samples does not more significant than in-distribution (ID) samples. In particular, the AUC-ROCs of all examined models below 0.50, which is in line with the definition of aleatoric uncertainty and the empirical evidence from \cite{kendalluncertainties}.

In our main manuscript, we demonstrate that the epistemic uncertainty of the ENet and MT-ENet is high with respect to the OOD datasets. In contrast, this aleatoric uncertainty experiment shows that the aleatoric uncertainty is not increased for the OOD datasets. This difference can support the claim of the evidential deep learning literature \cite{malinin2018predictive, sensoy2018evidential, gurevich2020gradient, malinin2020regression, amini2020deep} that the ENet can distinguish the aleatoric uncertainty and epistemic uncertainty. Furthermore, the results show that the additional MSE-based losses do not harm uncertainty disentangling.

\end{document}